\documentclass[journal]{IEEEtran}
\usepackage{latexsym}

\usepackage{enumerate}
\usepackage{graphicx}
\usepackage{subfigure}
\usepackage{multirow}

\usepackage{framed,multirow}
\usepackage{times}
\usepackage{graphicx} 
\usepackage{subfigure}
\usepackage{amssymb}
\usepackage{latexsym}

\usepackage{algorithm}
\usepackage{algorithmic}

\usepackage{hyperref}

\usepackage{url}
\usepackage{xcolor}
\usepackage{amsmath}
\usepackage{amsxtra}
\usepackage{braket}
\usepackage{array}
\usepackage{color}
\usepackage{url}
\usepackage{placeins}
\usepackage{threeparttable}

\usepackage{times}
\usepackage{balance}
\usepackage{enumerate}
\usepackage{multirow}
\usepackage{pifont}
\usepackage{marvosym}
\usepackage{ifsym}

\begin{document}
\title{Learning Backtrackless Aligned-Spatial Graph Convolutional Networks for Graph Classification}
\author{Lu~Bai,~\IEEEmembership{}Lixin~Cui,~\IEEEmembership{}Yuhang Jiao,~\IEEEmembership{}Luca~Rossi,~\IEEEmembership{}Edwin~R.~Hancock,~\IEEEmembership{IEEE~Fellow}

\thanks{Lu Bai (bailucs@cufe.edu.cn), Lixin Cui (Corresponding Author: cuilixin@cufe.edu.cn), Yuhang Jiao are with Central University of Finance and Economics, Beijing, China. Luca Rossi is with Southern University of Science and Technology, Shenzhen, China. Edwin R. Hancock is with University of York, York, UK.}
}

\markboth{Journal of \LaTeX\ Class Files,~Vol.~6, No.~1, January~2007}%
{Shell \MakeLowercase{\textit{et al.}}: Bare Demo of IEEEtran.cls
for Journals}
\maketitle

\begin{abstract}
In this paper, we develop a novel Backtrackless Aligned-Spatial Graph Convolutional Network (BASGCN) model to learn effective features for graph classification. Our idea is to transform arbitrary-sized graphs into fixed-sized backtrackless aligned grid structures and define a new spatial graph convolution operation associated with the grid structures. We show that the proposed BASGCN model not only reduces the problems of information loss and imprecise information representation arising in existing spatially-based Graph Convolutional Network (GCN) models, but also bridges the theoretical gap between traditional Convolutional Neural Network (CNN) models and spatially-based GCN models. Furthermore, the proposed BASGCN model can both adaptively discriminate the importance between specified vertices during the convolution process and reduce the notorious tottering problem of existing spatially-based GCNs related to the Weisfeiler-Lehman algorithm, explaining the effectiveness of the proposed model. Experiments on standard graph datasets demonstrate the effectiveness of the proposed model.
\end{abstract}


\IEEEpeerreviewmaketitle

\section{Introduction}\label{s1}

\IEEEPARstart{G}raph based representations are powerful tools to model complex systems that involve data lying on non-Euclidean spaces and that are naturally described in terms of relations between their components~\cite{DBLP:journals/tnn/ZambonAL18}, ranging from chemical compounds~\cite{DBLP:conf/icml/KriegeM12} to point clouds~\cite{DBLP:conf/cvpr/HarchaouiB07} and social networks~\cite{DBLP:conf/kdd/WangC016}. One fundamental challenge arising in the analysis of graph-based data is how to convert graph structures into numeric representations where standard machine learning techniques can be directly employed for graph classification or clustering. The aim of this paper is to develop a new Graph Convolutional Network (GCN) model to learn effective features for graph classification. Our idea is to transform arbitrary-sized graphs into fixed-sized backtrackless aligned grid structures and define a new backtrackless spatial graph convolution operation associated with the grid structures. We show that the proposed model not only bridges the theoretical gap between traditional Convolutional Neural Network (CNN) models and spatially-based GCN models, but also significantly reduces the notorious tottering problem of existing spatially-based GCNs related to the Weisfeiler-Lehman algorithm.

\subsection{Literature Review}
Broadly speaking, in the last three decades most classical state-of-the-art methods for the analysis of graph structures
can be separated into two categories, namely a) graph embedding methods and b) graph kernels. Approaches falling in the first category aim to convert graphs into elements of a vectorial space~\cite{DBLP:journals/pr/GibertVB12,DBLP:journals/pami/WilsonHL05,DBLP:conf/icml/KondorB08,DBLP:journals/tnn/RenWH11} where standard machine learning algorithms can be directly employed for graph data analysis. Unfortunately, these embedding methods tend to approximate structural correlations of graphs in a low dimensional pattern space, leading to structural information loss. To overcome this shortcoming, the proponents of graph kernel approaches suggest to characterize graph structures in a high dimensional Hilbert space and thus better preserve the structural information~\cite{DBLP:journals/ijon/Bai19,shervashidze2010weisfeiler,DBLP:conf/icml/KriegeM12,DBLP:journals/tnn/MartinoNS18,DBLP:conf/icml/Bai0ZH15,DBLP:conf/icdm/0001KM17}. One common limitation shared by both graph embedding methods and kernels is that of ignoring information from multiple
graphs. This is because graph embedding methods usually capture structural features of individual graphs, while graph
kernels reflect structural characteristics for pairs of graphs. Furthermore, since the process of computing the structural characteristics are separate from the classifier, both the graph embedding and kernel methods cannot provide an end-to-end learning architecture that simultaneously integrates the processes of graph characteristics learning and graph classification. In summary, these drawbacks influence the effectiveness of employing these traditional methods on graph classification tasks.

In recent years, due to the tremendous successes of deep learning networks in machine learning, there has been an increasing interest to generalize deep Convolutional Neural Networks (CNN)~\cite{DBLP:journals/cacm/KrizhevskySH17,DBLP:journals/pami/ZhangZHS16,DBLP:journals/pami/DongLHT16,DBLP:journals/pami/RenHG017,DBLP:conf/icml/YouYL19} to the graph domain. These novel deep learning networks on graphs are the so-called Graph Convolutional Networks (GCN)~\cite{DBLP:journals/corr/KipfW16} and have proven to be an effective way to extract highly meaningful statistical features for graph classification~\cite{DBLP:conf/nips/DefferrardBV16}. Generally speaking, most existing state-of-the-art GCN approaches can be divided into two main categories, i.e., GCN models based on a) spectral and b) spatial strategies. Specifically, approaches based on the spectral strategy define a convolution operation based on spectral graph theory~\cite{DBLP:journals/corr/BrunaZSL13,DBLP:journals/corr/HenaffBL15,DBLP:conf/nips/RippelSA15}. By transforming the graph into the spectral domain through the eigenvectors of the Laplacian matrix, these methods perform the filter operation by multiplying the graph by a series of filter coefficients. For instance, Bruna et al.~\cite{DBLP:journals/corr/BrunaZSL13} have developed a graph convolution network by defining a spectral filter based on computing the eigen-decomposition of the graph Laplacian matrix. To overcome the expensive computational complexity of the eigen-decomposition, Defferrard et al.~\cite{DBLP:conf/nips/DefferrardBV16} have approximated the spectral filters based on the Chebyshev expansion of the graph Laplacian. Unfortunately, most of the spectral-based approaches cannot be performed on graphs with different number of vertices and Fourier bases. Thus, these approaches work on same-sized graph structures and are usually employed for vertex classification tasks.

On the other hand, approaches based on the spatial strategy are not restricted to same-sized graph structures. These approaches generalize the graph convolution operation to the spatial structure of a graph by directly defining an operation on neighboring vertices~\cite{DBLP:conf/nips/AtwoodT16,DBLP:conf/nips/DuvenaudMABHAA15,DBLP:journals/corr/VialatteGM16}. For example, Duvenaud et al.~\cite{DBLP:conf/nips/DuvenaudMABHAA15} have proposed a spatially-based GCN model by defining a spatial graph convolution operation on the $1$-layer neighboring vertices to simulate the traditional circular fingerprint. Atwood and Towsley~\cite{DBLP:conf/nips/AtwoodT16} have proposed a spatially-based GCN model by performing spatial graph convolution operations on different layers of neighboring vertices rooted at a vertex. Although these spatially-based GCN models can be directly applied to real-world graph classification problems, they still need to further transform the multi-scale features learned from graph convolution layers into fixed-sized representations, so that the standard classifiers can be directly adopted for classifications. One way to achieve this is to directly sum up the learned local-level vertex features from the graph convolution operation as global-level graph features through a SumPooling layer. Since it is difficult to learn rich local vertex topological information from the global features, these spatially-based GCN methods associated with SumPooling have relatively poor performance on graph classification.

To overcome the above shortcoming of existing spatially-based GCN models, Zhang et al.~\cite{DBLP:conf/aaai/ZhangCNC18} have developed a novel spatially-based Deep Graph Convolutional Neural Network (DGCNN) model to preserve more vertex information. Specifically, they propose a new SortPooling layer to transform the extracted vertex features of unordered vertices from the spatial graph convolution layers into a fixed-sized local-level vertex grid structure. This is done by sequentially preserving a specified number of vertices with prior orders. With the fixed-sized grid structures of graphs to hand, a traditional CNN model followed by a Softmax layer can be directly employed for graph classification. Nieper et al.~\cite{DBLP:conf/icml/NiepertAK16}, on the other hand, have developed a different spatially-based Patchy-San Graph Convolutional Neural Network (PSGCNN) model to capture more vertex information through local neighbor vertices. Specifically, they extract and normalize a fixed-sized local neighborhood rooted at each vertex, where the vertices of each neighborhood are re-ordered based on the same graph labeling method and graph canonization tool. Since the normalized neighborhood can serve as the receptive field of its root vertex for the convolutional operation, this procedure naturally forms a local-level fixed-sized vertex grid structure for each graph. Thus, the graph convolution operation can be performed by sliding a fixed-sized classical standard convolutional filter over the neighboring vertices, i.e., the convolutional operation is similar to that performed on images with standard convolutional neural networks.

Although both the spatially-based DGCNN and PSGCNN models can capture rich graph characteristics residing on local-level vertices and outperform state-of-the-art GCN models on graph classification tasks, these methods establish the vertex order based on each individual graph. Thus, they cannot accurately reflect the topological correspondence information between graph structures. Moreover, both models lead to significant information loss, since those vertices associated with a lower ranking may be discarded. Finally, it has been shown in~\cite{DBLP:conf/aaai/ZhangCNC18} that most existing spatially-based GCN models~\cite{DBLP:conf/nips/AtwoodT16,DBLP:conf/nips/DuvenaudMABHAA15,DBLP:journals/corr/VialatteGM16} are related to the classical Weisfeiler-Lehman (WL) algorithm~\cite{shervashidze2010weisfeiler,DBLP:conf/icdm/0001KM17}. This is because the required convolution operation of these GCN models relies on aggregating the features of each vertex as well as its neighboring vertices, in a process that is similar to the WL algorithm, which propagates the features between each vertex and its neighboring vertices. Thus, similarly to the classical WL algorithm, these GCN models may also suffer from the well-known tottering problem~\cite{DBLP:conf/icml/Bai0ZH15}. In other words, these GCN models may propagate the feature information from the starting vertex to a second vertex and then immediately propagate the information back to the starting vertex, resulting in the creation of redundant feature information. 

\begin{figure*}
 \vspace{-10pt}
 \centering
\includegraphics[width=1\linewidth]{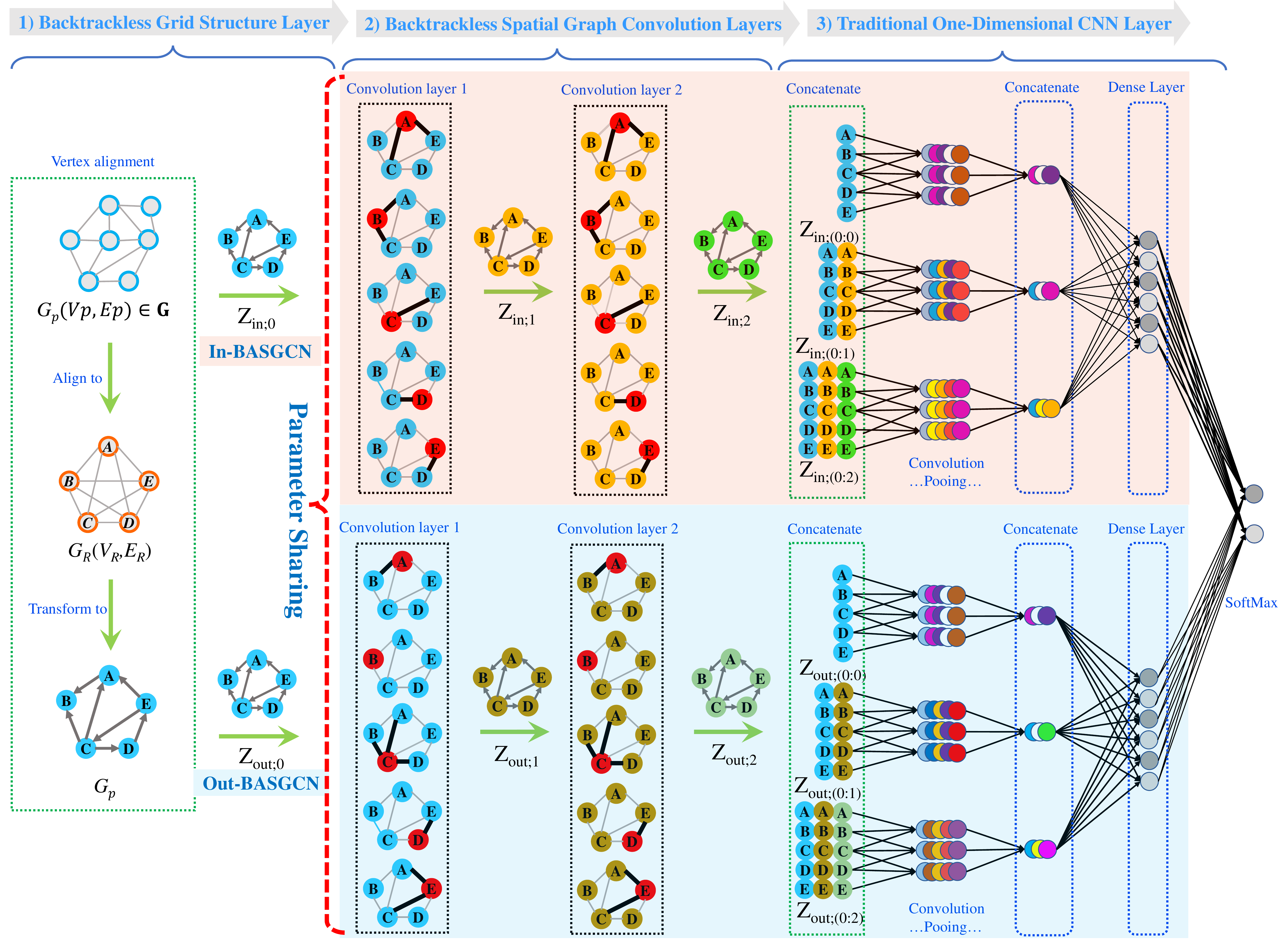}
 \vspace{-15pt}
  \caption{The architecture of the proposed BASGCN model (vertex features are visualized as different colors). An input graph $G_p(V_p,E_p)\in \mathbf{G}$ of arbitrary size is first aligned to the prototype graph $G_R(V_R,E_R)$. Then, $G_p$ is mapped into a fixed-sized backtrackless aligned vertex grid structure, where the vertex order follows that of $G_R$ and the associated aligned vertex adjacency matrix corresponds to a directed line graph, i.e., the connection between a pair of vertices is a directed edge. The grid structure of $G_p$ is passed through a pair of parallel stacked spatial graph convolution layers to extract multi-scale vertex features (i.e., $Z_{\mathrm{in};0}$ and $Z_{\mathrm{out};0}$ are the same), where the vertex information is propagated between specified vertices associated with the directed adjacency matrix. More formally, for each rooted vertex the upper convolution layers focus on aggregating the vertex features of the vertex itself as well as its in-neighbors (i.e., the vertices having directed edges to the rooted vertex), while the lower convolution layers focus on aggregating vertex features of the vertex itself as well as its out-neighbors (i.e., the vertices having directed edges from the rooted vertex to themselves). Note that both the upper and lower graph convolution layers share the same trainable parameters. \textbf{In the process of vertex information aggregation, the information is propagated along the directed edges, thus the information will not be immediately propagated back to the starting vertex, restricting the tottering problem}. Moreover, since the graph convolution layers preserve the original vertex order of the input grid structure, the concatenated vertex features through the graph convolution layers form a new vertex grid structure for $G_p$. This vertex grid structure is then passed to a traditional CNN layer for classification.}\label{f:QSGCNN}
 \vspace{-10pt}
\end{figure*}

\subsection{Contributions}

The aim of this paper is to address the shortcomings of existing methods by developing a novel Backtrackless Aligned-Spatial Graph Convolutional Network (BASGCN) model for graph classification tasks. To this end, we develop our recent work in~\cite{DBLP:conf/pkdd/Bai0BH19} one step further and generalize the original Aligned-Spatial GCN (ASGCN) model~\cite{DBLP:conf/pkdd/Bai0BH19} to a new backtrackless GCN model which reduces the aforementioned tottering problem. One key innovation of the new model is that of transitively aligning vertices between graphs. That is, given three vertices $v$, $w$ and $x$ from three different sample graphs, if $v$ and $x$ are aligned, and $w$ and $x$ are aligned, the proposed model can guarantee that $v$ and $w$ are also aligned. More specifically, similarly to the original ASGCN model, the proposed BASGCN model employs the transitive alignment procedure to transform arbitrary-sized graphs into fixed-sized aligned grid structures with consistent vertex orders, \textbf{guaranteeing that the vertices on the same spatial position are also transitively aligned to each other in terms of the topological structures}.

Since the process of constructing the grid structure does not discard any vertex, the proposed BASGCN model preserves the advantage of the original ASGCN model, i.e., it reduces the problems of information loss and imprecise information representation arising in existing spatially-based GCNs associated with SortPooling or SumPooling layers~\cite{DBLP:conf/nips/DefferrardBV16,DBLP:conf/aaai/ZhangCNC18}. Furthermore, the aligned grid structure of the proposed BASGCN model is a kind of backtrackless grid structure, i.e., it corresponds to a directed line graph rather than an undirected graph as in the original ASGCN model. Since the spatial graph convolution operation propagates the vertex feature information along the edges, the nature of the backtrackless grid implies that the information cannot be immediately propagated back to the starting vertex. Thus, this backtrackless structure provides a natural way to define a novel backtrackless spatial graph convolution operation that restricts the well-known tottering problem of existing spatially-based GCNs related to the WL algorithm~\cite{shervashidze2010weisfeiler}. As a result, the new proposed BASGCN model not only inherits all the advantages of the original ASGCN model, but also \textbf{further generalizes the original model to a new backtrackless GCN model that reduces the tottering problem and thus reflects richer graph characteristics} (see details in Sec.\ref{s4c}). The conceptual framework of the proposed BASGCN model is shown in Fig.\ref{f:QSGCNN}. Specifically, the main contributions of this work are threefold.

\textbf{First}, we introduce a new transitive vertex alignment method to map different arbitrary-sized graphs into fixed-sized backtrackless aligned grid structures, i.e., the aligned vertex grid structure as well as the associated backtrackless aligned vertex adjacency matrix. We show that the grid structures not only establish reliable vertex correspondence information between graphs, but also minimize the loss of structural information from the original graphs. Moreover, since the associated grid structure corresponds to a directed line graph, it provides a natural backtrackless structure to restrict the tottering problem.

\textbf{Second}, we develop a novel backtrackless spatially-based graph convolution model, i.e., the BASGCN model, for graph classification. More specifically, we propose a new backtrackless spatial graph convolution operation to extract multi-scale local-level vertex features. Unlike most existing spatially-based GCN models~\cite{DBLP:conf/nips/AtwoodT16,DBLP:conf/nips/DuvenaudMABHAA15,DBLP:journals/corr/VialatteGM16,DBLP:conf/aaai/ZhangCNC18} as well as the ASGCN model~\cite{DBLP:conf/pkdd/Bai0BH19}, which propagate features between vertices through the original vertex adjacency matrix or the undirected aligned vertex adjacency matrix, the proposed graph convolution layer propagates the feature information between aligned grid vertices through the associated backtrackless adjacency matrix. Since the backtrackless adjacency matrix corresponds to a directed line graph and provides a natural backtrackless structure, the proposed graph convolution operation can significantly restrict the tottering problem of most existing spatially-based GCNs as well as the original ASGCN model. Moreover, we show that the proposed convolution operation not only reduces the problems of information loss and imprecise information representation arising in existing spatially-based GCN models associated with SortPooling or SumPooling, but also theoretically relates to the classical convolution operation on standard grid structures. Thus, the proposed BASGCN model bridges the theoretical gap between traditional CNN models and spatially-based GCN models, and can adaptively discriminate the importance between specified vertices during the process of spatial graph convolution operations. Finally, since our backtrackless spatial graph convolution operation does not change the original spatial sequence of vertices, the proposed BASGCN model utilizes the traditional CNN to further learn graph features. In this way, we provide an end-to-end deep learning architecture that integrates the graph representation learning into both the backtrackless spatial graph convolutional layer and the traditional convolution layer for graph classification.

\textbf{Third}, we empirically evaluate the performance of the proposed BASGCN model on graph classification tasks. Experiments on widely used benchmarks demonstrate the effectiveness of the proposed method, when compared to state-of-the-art methods.

\subsection{Paper Outline}

The remainder of this paper is organized as follows. Section~\ref{s2} briefly reviews the existing spatially-based GCN models. Section~\ref{s3} introduces how to transform different arbitrary-sized graphs into fixed-sized backtrackless aligned grid structures. Section~\ref{s4} details the concept of the proposed BASGCN model. Section~\ref{s5} provides the experimental evaluation of the new method. Section~\ref{s6} concludes this work.
\section{Related Works of Spatially-based GCN Models}\label{s2}

In this section, we briefly review state-of-the art spatially-based GCN models in the literature. More specifically, we introduce the associated spatial graph convolution operation of the existing spatially-based Deep Graph Convolutional Neural Network (DGCNN) model~\cite{DBLP:conf/aaai/ZhangCNC18}. We refer this DGCNN model as a representative approach to analyze the common drawbacks arising in most existing spatially-based GCN models. To commence, consider a sample graph $G$ with $n$ vertices,  $X=(x_1,x_2,...,x_n)\in \mathbb{R}^{n\times c}$ is the collection of $n$ vertex feature vectors of $G$ in $c$ dimensions, and $A\in \mathbb{R}^{n\times n}$ is the vertex adjacency matrix ($A$ can be a weighted adjacency matrix). The spatial graph convolution operation of the DGCNN model takes the following form
\begin{equation}
Z= \mathrm{f}(\tilde{D}^{-1} \tilde{A} X W),\label{GCN_EAM}
\end{equation}
where $\tilde{A}=A+I$ is the adjacency matrix of graph $G$ with added self-loops, $\tilde{D}$ is the degree matrix of $\tilde{A}$ with $\tilde{A}_{[i,i]}=\sum_j \tilde{A}_{[i,j]}$, $W\in \mathbb{R}^{c\times c^{'}}$ is the matrix of trainable graph convolution parameters, $\mathrm{f}$ is a nonlinear activation function, and $Z\in \mathbb{R}^{n\times c^{'}}$ is the output of the convolution operation.

For the spatial graph convolution operation defined by Eq.(\ref{GCN_EAM}), the process $X W$ first maps the $c$-dimensional features of each vertex into a set of new $c^{'}$-dimensional features. Here, the filter weights $W$ are shared by all vertices. Moreover, $\tilde{A} Y$ ($Y:= {X} W$) aggregates the feature information of each vertex to its neighboring vertices as well as the vertex itself. The $i$-th row $(\tilde{A} Y)_{[i,:]}$ represents the extracted features of the $i$-th vertex, and corresponds to the summation or aggregation of $Y_{[i,:]}$ itself and $Y_{[j,:]}$ from its neighbor vertices. Multiplying by the inverse of $\tilde{D}$ (i.e., $\tilde{D}^{-1}$) can be seen as the process of normalizing and assigning equal weights between the $i$-th vertex and each of its neighbours.

Although the DGCNN model associated with convolution operation defined by Eq.(\ref{GCN_EAM}) has been proven a powerful GCN model for graph classification, it still suffers from the following two common drawbacks that arise in most existing spatially-based GCN models~\cite{DBLP:conf/nips/AtwoodT16,DBLP:conf/nips/DuvenaudMABHAA15,DBLP:journals/corr/VialatteGM16,DBLP:conf/aaai/ZhangCNC18}.\\

\noindent\textbf{Remark (Less Discrimination between Vertices):} Eq.(\ref{GCN_EAM}) indicates that the spatial graph convolution operation of the DGCNN model cannot discriminate the importance between specified vertices in the convolution operation process. This is because the required filter weights $W$ are shared by each vertex, i.e., the feature transformations of the vertices are all based on the same trainable function. Thus, the DGCNN model cannot directly influence the aggregation process of the vertex features. In fact, this problem also arises in other spatially-based GCN models that utilize the adjacency matrix for vertex information propagation, e.g., the Neural Graph Fingerprint Network (NGFN) model~\cite{DBLP:conf/nips/DuvenaudMABHAA15}, the Diffusion Convolution Neural Network (DCNN) model~\cite{DBLP:conf/nips/AtwoodT16}, etc. Since the associated spatial graph convolution operations of these models also take the similar form with that of the DGCNN model, i.e., the trainable parameters of their spatial graph convolution operations are also shared by each vertex. This drawback influences the effectiveness of the existing spatially-based GCN models for graph classification. \hfill$\Box$\\

\noindent\textbf{Remark (Tottering Problems between Vertices):} Zhang et al.~\cite{DBLP:conf/aaai/ZhangCNC18} have indicated the theoretical relationship between the DGCNN model and the classical WL algorithm~\cite{shervashidze2010weisfeiler}. The key idea of the WL method is to concatenate a vertex label with the labels of its neighboring vertices, and then sort the concatenated label lexicographically to assign each vertex a new label. The procedure repeats until a maximum iteration $h$, and each vertex label at an iteration $h$ corresponds to a subtree of height $h$ rooted at the vertex. If the concatenated label of two vertices are the same, the subtree rooted at the two vertices are isomorphic. To exhibit the relationship between the associated graph convolution operation of the DGCNN model defined by Eq.(\ref{GCN_EAM}) and the WL algorithm, we decompose Eq.(\ref{GCN_EAM}) into a row-wise manner, i.e.,
\begin{equation}
Z_{[i,:]}= \mathrm{Relu}({[\tilde{D}^{-1} \tilde{A}]}_{[i,:]} Y)= \mathrm{Relu} [\tilde{D}^{-1}_{[i,i]}(Y_{[i,:]}+ \sum_{j\in \Gamma(i)}Y_{[j,:]})],\label{GCN_WL}
\end{equation}
where $Y:={X} W$ and $\Gamma(i)$ corresponds to the set of neighboring vertices of the $i$-th vertices. For Eq.(\ref{GCN_WL}), $Y_{[i,:]}$ can be seen as the continuous valued vectorial vertex label of the $i$-th vertex. In a manner similar to the WL method, for each $i$-th vertex and its associated continuous label $Y_{[i,:]}$ Eq.(\ref{GCN_WL}) needs to propagate the continuous labels $Y_{[j,:]}$ of its neighboring vertices to its original label $Y_{[j,:]}$ as its new signature vector $\tilde{D}^{-1}_{[i,i]}(Y_{[i,:]}+ \sum_{j\in \Gamma(i)}Y_{[j,:]})$. The $\mathrm{Relu}$ function maps $\tilde{D}^{-1}_{[i,i]}(Y_{[i,:]}+ \sum_{j\in \Gamma(i)}Y_{[j,:]})$ to a new continuous vectorial label. As a result, the graph convolution operation defined by Eq.(\ref{GCN_WL}) can be seen as a soft version of the original WL algorithm, explaining the effectiveness of the DGCNN model. Unfortunately, similar to the classical WL algorithm, the DGCNN model also suffers from the tottering problem arising in the WL algorithm~\cite{DBLP:conf/icml/Bai0ZH15}. This is because, like the WL algorithm, the DGCNN model may propagate the feature information from the starting vertex to a vertex at the current convolution layer and then immediately propagate the information back to the starting vertex at the next convolution layer, resulting in redundant feature information. In fact, this problem also arises in other spatially-based GCN models~\cite{DBLP:conf/nips/AtwoodT16,DBLP:conf/nips/DuvenaudMABHAA15}, influencing their performance.\hfill$\Box$


\section{Constructing Aligned Backtrackless Grid Structures for Arbitrary Graphs}\label{s3}

Although, spatially-based GCN models are not restricted to the same graph structure, and can thus be applied for graph classification tasks. These methods still need to further transform the extracted multi-scale features from graph convolution layers into the fixed-sized characteristics through a SumPooling or SortPooling, so that the standard classifiers (e.g., the traditional convolutional neural network followed by a Softmax layer) can be directly employed for classifications. Unfortunately, these pooling operations usually cause information loss. In this section, we develop a transitive matching method to map different graphs of arbitrary sizes into fixed-sized backtrackless aligned grid structures, that can be directly utilized by the spatial graph convolution operation. Moreover, we show that the proposed grid structure not only integrates precise structural correspondence information but also minimizes the loss of structural information. Finally, we show that the proposed grid structure can provide a natural backtrackless structure to reduce the tottering problem arising in existing spatially-based GCN models.

\subsection{Identifying Transitive Vertex Alignment Information}
We introduce a new graph matching method to transitively align graph vertices. We first designate a family of prototype representations that encapsulate the principle characteristics over all vectorial vertex representations in a set of graphs $\mathbf{G}$. Assume there are $n$ vertices from all graphs in $\mathbf{G}$, and their associated $K$-dimensional vectorial representations are $\mathbf{{R}}^K =\{\mathrm{R}_1^K,\mathrm{R}_2^K,\ldots,\mathrm{R}_n^K\}$. We utilize $k$-means~\cite{witten2011data} to locate $M$ centroids over $\mathbf{{R}}^K$, by minimizing the objective function
\begin{equation}
\arg\min_{\Omega}  \sum_{j=1}^M \sum_{\mathrm{R}_i^K \in c_j} \|\mathrm{R}_i^K- \mu_j^K\|^2,\label{kmeans}
\end{equation}
where $\Omega=(c_1,c_2,\ldots,c_M)$ represents $M$ clusters, and $\mu_j^K$ the mean of the vertex representations belonging to the $j$-th cluster $c_j$.

Assume $\mathbf{G}=\{G_1,\cdots,G_p,\cdots,G_N\}$ is the graph sample set, where $G_p(V_p,E_p)\in {\mathbf{G}}$ is a sample graph of $\mathbf{G}$. For $G_p(V_p,E_p)$ and each vertex $v_i\in V_p$ associated with its $K$-dimensional vectorial representation $\mathrm{{R}}_{p;i}^K$, we initiate by locating a family of $K$-dimensional prototype representations as $\mathbf{PR}^K=\{\mu_1^K,\ldots,\mu_j^K,\ldots,\mu_M^K \}$ for the graphs over $\mathbf{G}$. To establish transitive correspondence information between different graphs, we follow the alignment procedure introduced by Bai et al.~\cite{DBLP:conf/icml/Bai0ZH15} for point matching in a pattern space. More formally, we align the vectorial vertex representations of each graph $G_p$ to the family of prototype representations in $\mathbf{PR}^K$, by computing a $K$-level affinity matrix in terms of the Euclidean distances between the two sets of points, i.e.,
\begin{align}
A^K_p(i,j)=\|\mathrm{{R}}_{p;i}^K - \mu_j^K\|_2.\label{AffinityM}
\end{align}
where $A^K_p$ is a ${|V_p|}\times {M}$ matrix, and each element $A^K_p(i,j)$ corresponds to the value of the distance between $\mathrm{{R}}_{p;i}^K$ and $\mu_j^K\in \mathbf{PR}^K$. If the element $A^K_p(i,j)$ is the smallest one in row $i$, we say that the vectrial representation $\mathrm{{R}}_{p;i}^K$ of $v\in V_p$ is aligned to the $j$-th prototype representation $\mu_j^K\in \mathbf{PR}^K$, i.e., the vertex $v_i$ is aligned to the $j$-th prototype representation. Note that for each graph there may be multiple vertices aligned to the same prototype representation. We record the correspondence information using the $K$-level correspondence matrix
$C^K_p\in \{0,1\}^{|V_p|\times M}$
\begin{equation}
C^K_p(i,j)=\left\{
\begin{array}{cl}
1   & \small{\mathrm{if} \  A^K_p(i,j) \ \mathrm{is \ the \ smallest \ in \ row } \ i} \\
0   & \small{\mathrm{otherwise}}.
\end{array} \right.
\label{CoMatrix}
\end{equation}

For each pair of graphs $G_p\in \mathbf{G}$ and $G_q\in \mathbf{G}$, if their vertices $v_p$ and $v_q$ are aligned to the same prototype representation ${\mu}_j^K\in \mathrm{PR}^K$, we say that $v_p$ and $v_q$ are also aligned. Thus, \textbf{we identify the transitive correspondence information between all graphs in $\mathbf{G}$, by aligning their vertices to a common set of prototype representations}.\\

\noindent\textbf{Remark:} The alignment process is equivalent to assigning the vectorial representation $\mathrm{{R}}_{p;i}^K$ of each vertex $v_i\in V_p$ to the mean $\mu_j^K$ of the cluster $c_j$. Thus, the proposed alignment procedure can be seen as an optimization process that gradually minimizes the inner-vertex-cluster sum of squares over the vertices of all graphs through $k$-means, and can establish reliable vertex correspondence information over all graphs.\hfill$\Box$
\begin{figure*}
 \vspace{-10pt}
 \centering
\includegraphics[width=1.0\linewidth]{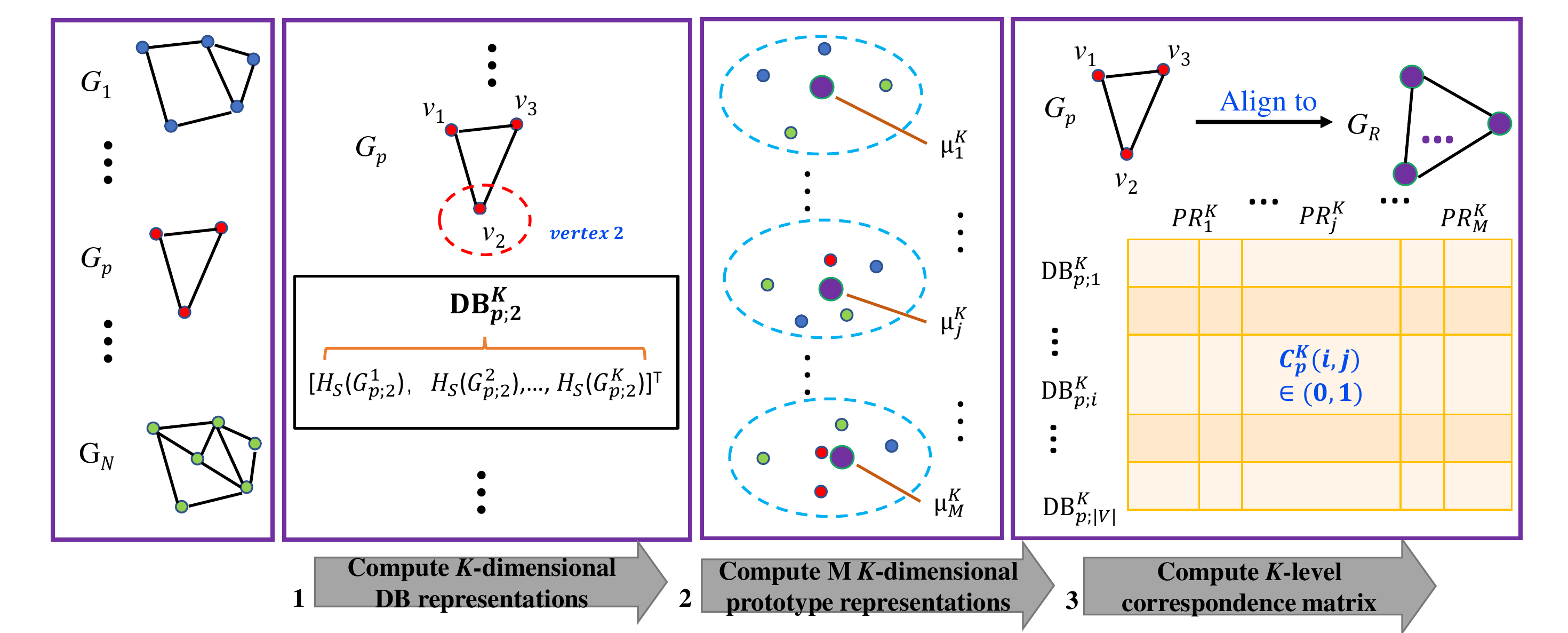}
 \vspace{-20pt}
  \caption{The procedure of computing the correspondence matrix. Given a set of graphs, for each graph $G_p$: (1) we compute the $K$-dimensional depth-based (DB) representation $\mathrm{{DB}}_{p;v}^K$ rooted at each vertex (e.g., vertex 2) as the $K$-dimensional vectorial vertex representation, where each element $H_s(G_{p;2}^K)$ represents the Shannon entropy of the $K$-layer expansion subgraph rooted at vertex $v_2$ of $G_p$~\cite{DBLP:journals/prl/Bai18}; (2) we identify a family of $K$-dimensional prototype representations $\mathbf{PR}^K = \{\mu_1^K,\ldots,\mu_j^K,\ldots,\mu_M^K \}$ using k-means on the $K$-dimensional DB representations of all graphs; (3) we align the $K$-dimensional DB representations to the $K$-dimensional prototype representations and compute a $K$-level correspondence matrix $C_p^K$.}\label{f:alignment}
 \vspace{-10pt}
\end{figure*}

\subsection{Aligned Grid Structures of Graphs}

We employ the transitive correspondence information to map arbitrary-sized graphs into fixed-sized backtrackless aligned grid structures, \textbf{i.e., the aligned vertex grid structures as well as the associated backtrackless aligned vertex adjacency matrices}. Assume $G_p(V_p,E_p,\tilde{A}_p)$ is a sample graph from the graph set $\mathbf{G}$, with $V_p$ representing the vertex set, $E_p$ representing the edge set, and $\bar{A}_p$ representing the vertex adjacency matrix with added self-loops (i.e., $\tilde{A}=A+I$, where $A$ is the original adjacency matrix with no self-loops and $I$ is the identity matrix). Let $X_p\in \mathbb{R}^{n\times c}$ be the collection of $n$ ($n=|V_p|$) vertex feature vectors of $G_p$ in $c$ dimensions. Note that, the row of $X_p$ follows the same vertex order of $\tilde{A}_p$. If $G_p$ are vertex attributed graphs, $X_p$ can be the one-hot encoding matrix of the vertex labels. For un-attributed graphs, we propose to use the vertex degree as the vertex label.

For each graph $G_p$, we utilize the proposed transitive vertex matching method to compute the $K$-level vertex correspondence matrix $C^K_p$ that records the correspondence information between the $K$-dimensional vectorial vertex representation of $G_p$ and the $K$-dimensional prototype representations in $\mathbf{PR}^K=\{\mu_1^K,\ldots,\mu_j^K,\ldots,\mu_M^K \}$. With $C^K_p$ to hand, we compute the $K$-level aligned vertex feature matrix for $G_p$ as
\begin{equation}
\bar{X}_{p}^{K}= (C^K_p)^T X_p,\label{alignDB}
\end{equation}
where $\bar{X}_{p}^{K} \in \mathbb{R}^{M\times c}$ and each row of $\bar{X}_{p}^{K}$ represents the feature of a corresponding aligned vertex. Moreover, we also compute the associated $K$-level aligned vertex adjacency matrix for $G_p$ as
\begin{equation}
\bar{A}_{p}^{K}= (C^K_p)^T (\tilde{A}_{p}) (C^K_p),\label{alignA}
\end{equation}
where $\bar{A}_{p}^{K}\in \mathbb{R}^{M\times M}$. Both $\bar{X}_{p}^{K}$ and $\bar{A}_{p}^{K}$ are indexed by the corresponding prototypes in $\mathbf{PR}^K$. Since $\bar{X}_{p}^{K}$ and $\bar{A}_{p}^{K}$ are computed from the original vertex feature matrix $X_p$ and the original adjacency matrix $\tilde{A}_p$, respectively, by mapping the original feature and adjacency information of each vertex $v_p\in V_p$ to that of the new aligned vertices, $\bar{X}_{p}^{K}$ and $\bar{A}_{p}^{K}$ encapsulate the original feature and structural information of $G_p$. Note that, according to Eq.~\ref{CoMatrix}, each prototype may be aligned by more than one vertex from $V_p$, $\bar{A}_{p}^{K}$ may be a weighted adjacency matrix.

In order to construct the fixed-sized aligned grid structure for each graph $G_p\in \mathbf{G}$, we need to sort the vertices to determine their spatial orders. Since the vertices of each graph are all aligned to the same prototype representations, we sort the vertices of each graph by reordering the prototype representations. To this end, we construct a prototype graph $G_{\mathrm{R}}(V_{\mathrm{R}},E_{\mathrm{R}})$ that captures the pairwise similarity between the $K$-dimensional prototype representations in $\mathbf{PR}^K$, with each vertex $v_j\in V_{\mathrm{R}}$ representing the prototype representation $\mu_j^K\in \mathbf{PR}^K$ and each edge $(v_j,v_k)\in E_{\mathrm{R}} $ representing the similarity between $\mu_j^K\in \mathbf{PR}^K$ and $\mu_k^K\in \mathbf{PR}^K$. The similarity between two vertices of $G_{\mathrm{R}}$ is computed as \begin{equation}
s(\mu_j^K,\mu_k^K)=\exp (-\frac{\| \mu_j^K-\mu_k^K   \|_2}{K}).
\end{equation}
The degree of each prototype representation $\mu_j^K$ is $D_R(\mu_j^K)=\sum_{k=1}^{M}s(\mu_j^K,\mu_k^K)$. We propose to sort the $K$-dimensional prototype representations in $\mathbf{PR}^K$ according to their degree $D_R(\mu_j^K)$. Then, we rearrange $\bar{X}_{p}^{K}$ and $\bar{A}_{p}^{K}$ accordingly.

To construct reliable grid structures for graphs, in this work we employ the depth-based (DB) representations as the vectorial vertex representations to compute the required $K$-level vertex correspondence matrix $C_p^K$. The DB representation of each vertex is defined by measuring the entropies on a family of $k$-layer expansion subgraphs rooted at the vertex~\cite{DBLP:journals/pr/BaiH14}, where the parameter $k$ varies from $1$ to $K$. It is shown that such a $K$-dimensional DB representation encapsulates rich entropy content flow from each local vertex to the global graph structure, as a function of depth. The process of computing the correspondence matrix $C_p^K$ associated with depth-based representations is shown in Fig.\ref{f:alignment}. When we vary the number of layers $K$ from $1$ to $L$ (i.e., $K\leq L$), we compute the final \textbf{aligned vertex grid structure} for each graph $G_p\in \mathbf{G}$ as
\begin{equation}
\bar{X}_{p}= \sum_{K=1}^L \frac{\bar{X}_{p}^{K}}{L},\label{AlignV}
\end{equation}
and the associated aligned grid vertex adjacency matrix as
\begin{equation}
\bar{A}_{p}= \sum_{K=1}^L \frac{\bar{A}_{p}^{K}}{L},\label{AlignA}
\end{equation}
where $\bar{X}_{p}\in \mathbb{R}^{M\times c}$, $\bar{A}_{p}\in \mathbb{R}^{M\times M}$, the $i$-th row of $\bar{X}_{p}$ corresponds to the feature vector of the $i$-th aligned grid vertex, and the $i$-row and $j$-column element of $\bar{A}_{p}$ corresponds to the adjacent information between the $i$-th and $j$-th grid vertices.

Note that, the adjacency matrix $\bar{A}_{p}$ corresponds to an undirected graph. Directly associating this matrix with existing spatial graph convolution operations may also suffer from tottering problems, since the vertex feature information may be propagated from the starting vertex to a vertex through an undirected edge (i.e., a bidirectional edge) and then immediately propagated back to the starting vertex through the same edge. This in turn results redundant feature information, and influences the performance of existing spatially-based GCN models. To address this problem, we propose to transform $\bar{A}_{p}$ into a backtrackless adjacency matrix $\bar{A}^D_{p}$, that corresponds to a directed line graph. More formally, with the undirected aligned vertex adjacency matrix $\bar{A}^D_{p}$ to hand, we commence by computing the degree of each $i$-th aligned grid vertex as $\bar{D}^D_{p}(i)=\sum_{j}\bar{A}^D_{p}(i,j)$. The probability of the classical steady state random walk visiting the $i$-th vertex is then computed as
\begin{equation}
P(i)=\bar{D}^D_{p}(i)/ \sum_j \bar{D}^D_{p}(j).
\end{equation}
We compute the \textbf{backtrackless aligned grid vertex adjacency matrix} $\bar{A}^D_{p}$ of each graph $G_p$ by replacing each bidirectional edge residing on $\bar{A}_{p}$ as a directed edge associated with the visiting probabilities of classical random walks, i.e.,
\begin{equation}
\bar{A}^D_{p}(v_i,v_j)=\left\{ \begin{array}{cl}
 \bar{A}_{p}(v_i,v_j) &\mbox{ if \ $P(i)\leqslant P(j)$, } \\
 0 &\mbox{ otherwise.}
       \end{array} \right.\label{AlignDA}.
\end{equation}
where $v_i$ and $v_j$ are the $i$-th and $j$-th aligned grid vertices, and $P(i)$ and $P(j)$ are the probabilities of the classical random walk visiting $v_i$ and $v_j$. Clearly, $\bar{A}^D_{p}$ corresponds to a directed line graph. Unlike the undirected grid vertex adjacency matrix $\bar{A}^D_{p}$, the backtrackless grid vertex adjacency matrix $\bar{A}^D_{p}$ is not a symmetric matrix. If the $i$-row and $j$-column element of $\bar{A}^D_{p}$ is greater than $0$, we say that there is a directed edge from the grid vertex $v_i$ to the grid vertex $v_j$. \textbf{Since, the vertex feature information cannot immediately propagate back to the starting vertex along a directed edge within the spatial graph convolution operation, $\bar{A}^D_{p}$ provides a natural backtrackless structure to restrict the tottering problem}. Finally, note that, Eq.(\ref{AlignDA}) will not discard the self-loop information residing on the trace of $\bar{A}_{p}$.\\

\noindent\textbf{Remark:} Eq.(\ref{AlignV}) and Eq.(\ref{AlignDA}) indicate that they can transform the original graph $G_p\in \mathbf{G}$ with arbitrary number of vertices $|V_p|$ into a new backtrackless aligned grid graph structure with the same number of vertices, where $\bar{X}_{p}$ is the corresponding aligned grid vertex feature matrix and $\bar{A}^D_{p}$ is the corresponding backtrackless aligned grid vertex adjacency matrix. Since both $\widehat{X}_{p}$ and $\bar{A}^D_{p}$ are mapped through the original graph $G_p$, they not only reflect reliable structure correspondence information between $G_p$ and the remaining graphs in graph set $\mathbf{G}$ but also encapsulate more original feature and structural information of $G_p$. Furthermore, since the orientation of each directed edge residing on the backtrackless adjacency matrix $\bar{A}^D_{p}$ is from a vertex with a lower visiting probability of random walks to that with a higher visiting probability of random walks, $\bar{A}^D_{p}$ encapsulates rich visiting information of random walks. \hfill$\Box$


\section{The Backtrackless Aligned-Spatial Graph Convolutional Network Model}\label{s4}

In this section, we propose a new spatially-based GCN model, i.e., the Backtrackless Aligned-Spatial Graph Convolutional Network (BASGCN) model. The core stage of a spatially-based GCN model is the associated graph convolution operation that extracts multi-scale features for each vertex based on the original features of its neighbour vertices as well as itself. As we have stated, most existing spatially-based GCN models perform the convolution operation by first applying a trainable parameter matrix to map the original feature of each vertex in $c$ dimensions to that in $c'$ dimensions, and then averaging the vertex features of specified vertices~\cite{DBLP:conf/nips/AtwoodT16,DBLP:conf/nips/DuvenaudMABHAA15,DBLP:journals/corr/VialatteGM16,DBLP:conf/aaai/ZhangCNC18}. Since the trainable parameter matrix is shared by all vertices, these models cannot discriminate the importance of different vertices and have inferior ability to aggregate vertex features. Moreover, as we have indicated, most existing spatially-based GCN models are theoretically related to the classical WL algorithm~\cite{shervashidze2010weisfeiler}, and the required convolution operation of these GCN models relies on the vertex feature propagation between each vertex and its neighboring vertices~\cite{DBLP:conf/nips/AtwoodT16,DBLP:conf/nips/DuvenaudMABHAA15,DBLP:journals/corr/VialatteGM16,DBLP:conf/aaai/ZhangCNC18}. Thus, similar to the WL algorithm, these WL analogous GCN models may propagate the feature information from the starting vertex to a vertex and then immediately propagate the information back to the starting vertex, resulting in redundant feature information. To overcome these shortcomings, in this section we first propose a new backtrackless spatial graph convolution operation associated with the backtrackless aligned grid structures of graphs. Unlike existing methods, the trainable parameters of the proposed convolution operation can directly influence the aggregation of the aligned grid vertex features, thus the proposed convolution operation can discriminate the importance between specified aligned grid vertices. Furthermore, since the process of the vertex feature information propagation relies on the backtrackless aligned grid vertex adjacency matrix, the proposed convolution operation can significantly reduce the tottering problem. Finally, we introduce the architecture of the BASGCN model associated with the proposed convolution operation.

\subsection{The Backtrackless Spatial Graph Convolution Operation}\label{SGCP}

In this subsection, we propose a new backtrackless spatial graph convolution operation to further extract multi-scale features of graphs, by propagating features between aligned grid vertices through the backtrackless aligned grid vertex adjacency matrix. Specifically, given a sample graph $G(V,E)$ with its aligned vertex grid structure $\bar{X}\in \mathbb{R}^{M\times c}$ and the associated backtrackless aligned grid vertex adjacency matrix $\bar{A}^D\in \mathbb{R}^{M\times M}$, the proposed backtrackless spatial graph convolution operation takes the following forms
\begin{equation}
Z^h_\mathrm{in}= \mathrm{Relu}( \bar{D}^{-1}_\mathrm{in} \bar{A}_\mathrm{in} \sum_{j=1}^c{(\bar{X} \odot W^h)}_{[:,j]}),\label{GCN_EQ}
\end{equation}
and
\begin{equation}
Z^h_\mathrm{out}= \mathrm{Relu}( \bar{D}^{-1}_\mathrm{out} \bar{A}_\mathrm{out} \sum_{j=1}^c{(\bar{X} \odot W^h)}_{[:,j]}),\label{GCN_EQout}
\end{equation}
where $\odot$ represents the element-wise Hadamard product, $\bar{A}_\mathrm{in}$ equals to ${(\bar{A}^D)}^T$ and is the \textbf{in-adjacency matrix} (i.e., for the $i$-th row, its $j$-th column elements of $\bar{A}_\mathrm{in}$ correspond to the directed edges to the $i$-th grid vertex from these $j$-th grid vertices, and we regard these $j$-th grid vertices as the \textbf{in-neighboring vertices} of the $i$-th grid vertex), $\bar{A}_\mathrm{out}$ equals to ${\bar{A}^D}$ and is the \textbf{out-adjacency matrix} (i.e., for the $i$-th row, its $j$-th column elements of $\bar{A}_\mathrm{out}$ correspond to the directed edges from the $i$-th grid vertex to these $j$-th grid vertices, and we regard these $j$-th grid vertices as the \textbf{out-neighboring vertices} of the $i$-th grid vertex), $\bar{D}_\mathrm{in}$ is the \textbf{in-degree matrix} of $\bar{A}_\mathrm{in}$, $\bar{D}_\mathrm{out}$ is the \textbf{out-degree matrix} of $\bar{A}_\mathrm{out}$. More specifically, Eq.(\ref{GCN_EQ}) corresponds to the \textbf{in-spatial graph convolution operation} (i.e., for each grid vertex this convolution operation focuses on propagating the feature information between itself and its in-neighboring vertices). Eq.(\ref{GCN_EQout}) corresponds to the \textbf{out-spatial graph convolution operation} (i.e., for each grid vertex this convolution operation focuses on propagating the feature information between itself and its out-neighboring vertices). The in-spatial and out-spatial convolution operations share the same trainable parameter matrix $W^h\in \mathbb{R}^{M\times c}$ for both their $h$-th convolution filters with the filter size $M\times 1$ and the channel number $c$. $\mathrm{Relu}$ is the rectified linear units function (i.e., a nonlinear activation function), and $Z_\mathrm{in}^h\in \mathbb{R}^{M\times 1}$ and $Z_\mathrm{out}^h\in \mathbb{R}^{M\times 1}$ are the output activation matrices for the in-spatial and out-spatial convolution operations.
\begin{figure*}
 \vspace{-10pt}
 \centering
\includegraphics[width=0.99\linewidth]{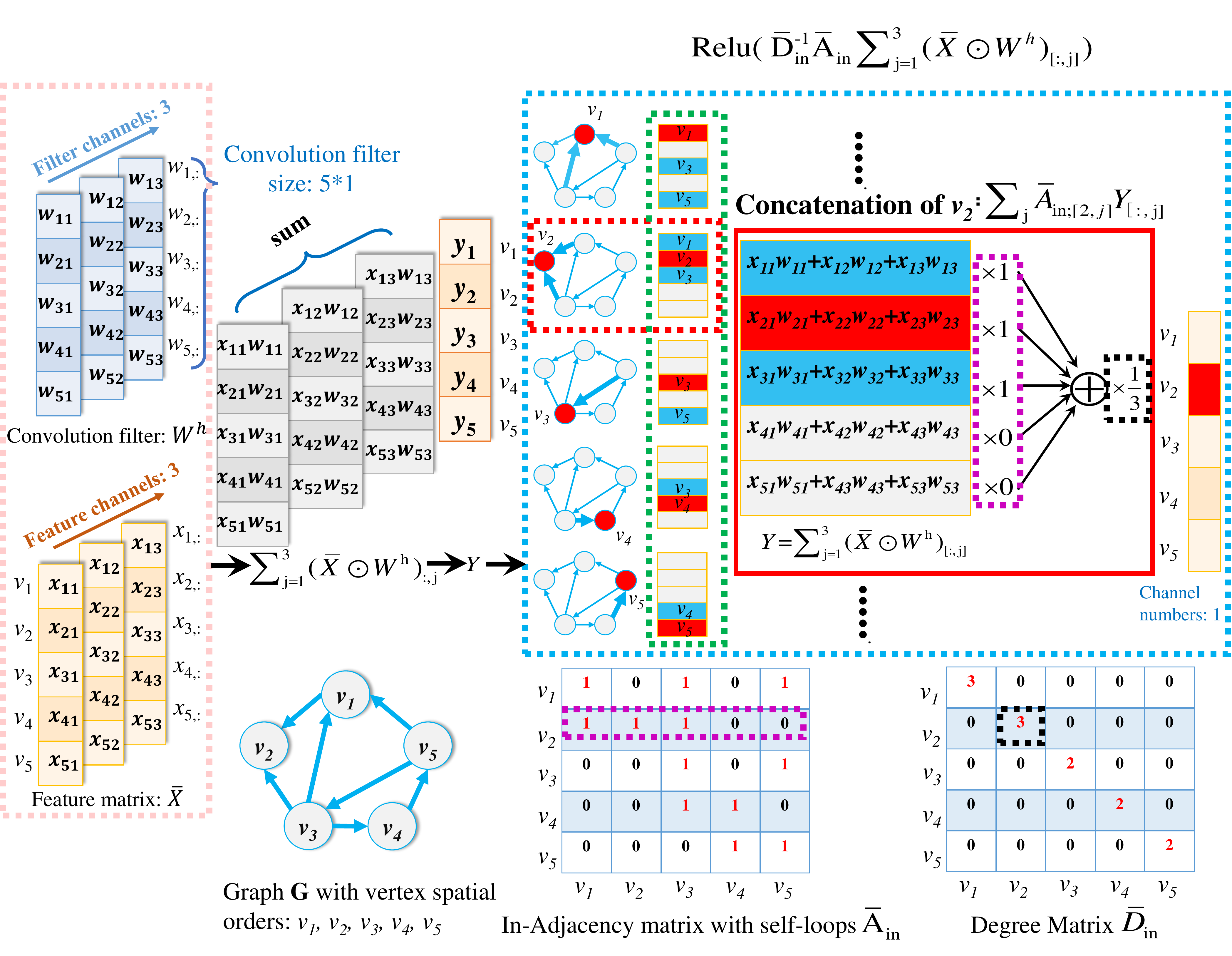}
 \vspace{-10pt}
  \caption{An Instance of the Proposed Backtrackless In-Spatial Graph Convolution Operation.}\label{f:alignment}
 \vspace{-10pt}
\end{figure*}

An instance of the proposed in-spatial graph convolution operation defined by Eq.(\ref{GCN_EQ}) is shown in Fig.\ref{f:alignment}. Specifically, this convolution operation consists of four steps. \textbf{In the first step}, the procedure $\sum_{j=1}^c{(\bar{X} \odot W^h)}_{[:,j]}$ commences by computing the element-wise Hadamard product between $\bar{X}$ and $W^h$, and then summing the channels of $\bar{X} \odot W^h$ (i.e., summing the columns of $\bar{X} \odot W^h$). Fig.\ref{f:alignment} exhibits this process. Assume $\bar{X}$ is the collection of $5$ aligned grid vertex feature vectors in the $3$ dimensions (i.e., $3$ feature channels), $W^h$ is the $h$-th convolution filter with the filter size $5\times 1$ and the channel number $3$. The resulting $\sum_{j=1}^3{(\bar{X} \odot W^h)}_{[:,j]}$ first assigns the feature vector $x_{[i,:]}$ of each $i$-th aligned grid vertex a different weighted vector $w_{[i,:]}$, and then sums the channels of each weighted feature vector. For the first step, $\sum_{j=1}^c{(\bar{X} \odot W^h)}_{[:,j]}$ can be seen as a new weighted aligned vertex grid structure with $1$ vertex feature channel. \textbf{The second step} $\bar{A}_\mathrm{in} Y$, where $Y:= \sum_{j=1}^c{(\bar{X} \odot W^h)}_{[:,j]}$, propagates the weighted feature information between each aligned grid vertex as well as its in-neighboring aligned grid vertices. Specifically, each $i$-th row $(\bar{A}_\mathrm{in} Y)_{[i,:]}$ of $\bar{A}_\mathrm{in} Y$ equals to $\sum_j \bar{A}_{\mathrm{in};{[i,j]}} Y_{[:,j]}$, and can be seen as the aggregated feature vector of the $i$-th aligned grid vertex by summing its original weighted feature vector as well as all the original weighted feature vectors of the $j$-th aligned grid vertex that has a directed edge to it (i.e., its in-neighboring vertices). Note that, since the first step has assigned each $i$-th aligned grid vertex a different weighted vector $w_{[i,:]}$, this aggregation procedure is similar to performing a standard fixed-sized convolution filter on a standard grid structure, where the filter first assigns different weighted vectors to the features of each grid element as well as its neighboring grid elements and then aggregates (i.e., sum) the weighted features as the new feature for each grid element. \textbf{This indicates that the trainable parameter matrix $W^h$ of the proposed convolution operation can directly influence the aggregation process of the vertex features, i.e., it can adaptively discriminate the importance between specified in-neighboring vertices.} Fig.\ref{f:alignment} exhibits this propagation process. For the $2$-nd aligned grid vertex $v_2$ (marked by the red broken-line frame), the $1$-st and $3$-rd aligned grid vertices $v_1$ and $v_3$ are its in-neighboring vertices. The process of computing $\sum_j \bar{A}_{\mathrm{in};{[2,j]}} Y_{[:,j]}$ (marked by the red real-line frame) aggregates the weighted feature vectors of aligned grid vertex $v_2$ as well as its in-neighboring aligned grid vertices $v_1$ and $v_3$ as the new feature vector of $v_2$. The vertices participating in this aggregation process are indicated by the $2$-nd row of $\bar{A}_\mathrm{in}$ (marked by the purple broken-line frame on $\bar{A}$) that encapsulates the in-adjacent information of aligned grid vertices. \textbf{The third step} normalizes each $i$-th row of $\bar{A}_\mathrm{in} Y$ by multiplying $\bar{D}_{\mathrm{in};{[i,i]}}^{-1}$, where $\bar{D}_{\mathrm{in};{[i,i]}}$ is the $i$-th diagonal element of the in-degree matrix $\bar{D}_\mathrm{in}$. This process can guarantee a fixed feature scale after the proposed convolution operation. Specifically, Fig.\ref{f:alignment} exhibits this normalization process. The aggregated feature of the $2$-nd aligned grid vertex (marked by the red real-line frame) is multiplied by $3^{-1}$, where $3$ is the $3$-rd diagonal element of $\bar{D}_\mathrm{in}$ (marked by the black broken-line frame on $\bar{D}_\mathrm{in}$). \textbf{The last step} employs the Relu activation function and outputs the convolution result. Note that, since the proposed in-spatial graph convolution operation defined by Eq.(\ref{GCN_EQ}) only extracts new features for the aligned grid vertex and does not change the orders of the aligned vertices, the output $Z_\mathrm{in}^h$ is still an aligned vertex grid structure with the same vertex order of $\bar{X}$.

Similar to the in-spatial graph convolution operation, the out-spatial graph convolution operation defined by Eq.(\ref{GCN_EQout}) can also be explained by Fig.\ref{f:alignment}, by replacing both the in-adjacency matrix $\bar{A}_\mathrm{in}$ and the in-degree matrix $\bar{D}_\mathrm{in}$ as the out-adjacency matrix $\bar{A}_\mathrm{out}$ and the out-degree matrix $\bar{D}_\mathrm{out}$. As a result, different from the in-spatial graph convolution operation, the out-spatial graph convolution operation focuses on propagating the weighted feature information between each aligned grid vertex as well as its out-neighboring aligned grid vertices. Moreover, the output $Z_\mathrm{out}^h$ of the out-spatial convolution operation is also an aligned vertex grid structure with the same vertex order of $\bar{X}$.

Finally, the above theoretical explanation indicates that both the in-spatial and out-spatial graph convolution operations can significantly reduce the drawback of tottering problems, that arises in mosting existing spatially-based GCN models~\cite{DBLP:conf/nips/AtwoodT16,DBLP:conf/nips/DuvenaudMABHAA15,DBLP:journals/corr/VialatteGM16} theoretically related to the WL algorithm~\cite{shervashidze2010weisfeiler}. This is because both the convolution operations are defined by propagating vertex feature information through the backtrackless aligned grid vertex adjacency matrix (i.e., the in and our adjacency matrices $\bar{A}_\mathrm{in}$ and $\bar{A}_\mathrm{out}$), that corresponds to a directed line graph. Thus, the vertex feature information cannot be immediately propagated back to the starting vertex through a directed edge within the spatial graph convolution operation. In other words, the current convolution operation can propagate the vertex feature information from a starting vertex to a vertex along (for the in-convolution) or against (for the out-convolution) a directed edge, but the next convolution operation cannot immediately propagate the information back from the vertex to the starting vertex against (for the in-convolution) or along (for the out-convolution) the same directed edge. As a result, \textbf{both the in-spatial and out-spatial graph convolution operations are Backtrackless Spatial Graph Convolution Operations}.

\subsection{The Architecture of the Proposed BASGCN Model}\label{s:arc}

We introduce the architecture of the proposed BASGCN model, that has been shown in Fig.\ref{f:QSGCNN}. Specifically, the architecture is composed of three sequential stages, i.e., 1) the backtrackless grid structure construction/input layer, 2) the backtrackless spatial graph convolution layer, and 3) the traditional One-dimensional CNN layers.\\

\noindent\textbf{The Backtrackless Grid Structure Construction/Input Layer:} For the proposed BASGCN model, we commence by employing the transitive vertex matching method defined earlier to convert each graph $G\in \mathbf{G}$ of arbitrary sizes into the fixed-sized backtrackless aligned grid structure, including the aligned vertex grid structure $\bar{X}$ as well as the associated backtrackless aligned grid vertex adjacency matrix $\bar{A}$. Then, we pass the grid structures to the proposed BASGCN model.\\

\noindent\textbf{The Spatial Graph Convolutional Layer:} For each graph $G$, to extract multi-scale features of the aligned grid vertices, we define a pair of paralleling stacked multiple backtrackless graph convolution layers associated with the proposed in-spatial and out-spatial graph convolution operations respectively, i.e., the \textbf{In-BASGCN} network focusing on aggregating vertex features of itself and its in-neighboring vertices and the \textbf{Out-BASGCN} network focusing on aggregating vertex features of itself and its out-neighboring vertices (see Fig.\ref{f:QSGCNN} for details). Both networks are backtrackless GCNs. More formally, the In-BASGCN network associated with Eq.(\ref{GCN_EQ}) and the Out-BASGCN network associated with Eq.(\ref{GCN_EQout}) are separately defined as
\begin{equation}
Z^h_{\mathrm{in};t}= \mathrm{Relu}( \bar{D}_{\mathrm{in}}^{-1} \bar{A} \sum_{j=1}^{H_{t-1}}{(Z_{\mathrm{in};t-1} \odot W^h_{t})}_{[:,j]}),\label{GCN_EQM}
\end{equation}
and
\begin{equation}
Z^h_{\mathrm{out};t}= \mathrm{Relu}( \bar{D}_{\mathrm{out}}^{-1} \bar{A} \sum_{j=1}^{H_{t-1}}{(Z_{\mathrm{out};t-1} \odot W^h_{t})}_{[:,j]}),\label{GCN_EQMout}
\end{equation}
where $Z_{\mathrm{in};0}$ and $Z_{\mathrm{out};0}$ are the same and equal to the aligned vertex grid structure $\bar{X}$, Eq.(\ref{GCN_EQM}) and Eq.(\ref{GCN_EQMout}) correspond to the stacked In-BASGCN and Out-BASGCN networks and share the same trainable parameters, $H_{t-1}$ is the number of convolution filters in the $(t-1)$-th graph convolution layer for the stacked In-BASGCN or Out-BASGCN network, $Z_{\mathrm{in};t-1} \in \mathbb{R}^{M\times H_{t-1}}$ and $Z_{\mathrm{out};t-1} \in \mathbb{R}^{M\times H_{t-1}}$ are the concatenated outputs of all the $H_{t-1}$ convolution filters in the $(t-1)$-th graph convolution layer of the stacked In-BASGCN and Out-BASGCN networks, $Z^h_{\mathrm{in};t}$ and $Z^h_{\mathrm{out};t}$ are the outputs of the $h$-th convolution filter in layer $t$ of the stacked In-BASGCN and Out-BASGCN networks, and $W^h_t \in \mathbb{R}^{M\times H_{t-1}}$ is the trainable parameter matrix of the $h$-th convolution filter in layer $t$ with the filter size $M \times 1$ and the channel number $H_{t-1}$ for the stacked In-BASGCN or Out-BASGCN networks.\\

\noindent\textbf{The Traditional One-dimensional CNN Layer:} For the In-BASGCN and Out-BASGCN networks, we horizontally concatenate the outputs $Z_{\mathrm{in};t}$ and $Z_{\mathrm{out};t}$ of their each $t$-th spatial graph convolution layers associated with the outputs of their previous $1$ to $(t-1)$-th spatial graph convolution layers as well as their original inputs $Z_{\mathrm{in};0}$ and $Z_{\mathrm{out};0}$ as $Z_{\mathrm{in};(0:t)}$ and $Z_{\mathrm{out};(0:t)}$, i.e., $Z_{\mathrm{in};(0:t)}=[Z_{\mathrm{in};0},Z_{\mathrm{in};1},\ldots,Z_{\mathrm{in};t}]$ and $Z_{\mathrm{in};0:t}\in  \mathbb{R}^{M\times (c+\sum_{z=1}^t H_t)}$, and $Z_{\mathrm{out};(0:t)}=[Z_{\mathrm{out};0},Z_{\mathrm{out};1},\ldots,Z_{\mathrm{out};t}]$ and $Z_{\mathrm{out};0:t}\in  \mathbb{R}^{M\times (c+\sum_{z=1}^t H_t)}$. As a result, for the concatenated outputs $Z_{\mathrm{in};0:t}$ and $Z_{\mathrm{out};0:t}$, each of their rows can be seen as the new multi-scale features for the corresponding aligned grid vertex. Since both $Z_{\mathrm{in};0:t}$ and $Z_{\mathrm{out};0:t}$ are still aligned vertex grid structures, one can directly utilize the traditional CNN on these grid structures. Specifically, Fig.\ref{f:QSGCNN} exhibits the architecture of a pair of paralleling traditional CNN layers, where the upper CNN layers are associated with each $Z_{\mathrm{in};0:t}$ of the In-BASGCN network, and the lower CNN layers are associated with each $Z_{\mathrm{out};0:t}$ of the Out-BASGCN network. Similar to the In-BASGCN and Out-BASGCN networks, both the upper and lower CNN layers also share the same trainable parameters. Moreover, each concatenated vertex grid structure $Z_{\mathrm{in};0:t}$ or $Z_{\mathrm{out};0:t}$ can be seen as a $M \times 1$ (in Fig.\ref{f:QSGCNN} $M = 5$) vertex grid structure and each vertex is represented by a $(c+\sum_{z=1}^t H_t)$-dimensional feature, i.e., the channel of each grid vertex is $c+\sum_{z=1}^t H_t$. Then, we add a one-dimensional convolution layer for each $Z_{\mathrm{in};0:t}$ or $Z_{\mathrm{out};0:t}$. The convolution operation can be performed by sliding a fixed-sized filter of size $k\times 1$ (in Fig.\ref{f:QSGCNN} $k = 3$) over the spatially neighboring vertices. After this, several AvgPooling layers and remaining one-dimensional convolution layers can be added to learn the local patterns on the aligned grid vertex sequence. Finally, when we vary $t$ from $0$ to $T$ (in Fig.\ref{f:QSGCNN} $T = 2$), we will obtain $T+1$ extracted pattern representations for the upper or lower CNN layers. We concatenate the extracted patterns of each $Z_{\mathrm{in};0:t}$ or $Z_{\mathrm{out};0:t}$ and add a fully-connected layer. A Softmax layer is added and follows the fully-connected layers of both the upper and lower CNN layers.

\subsection{Discussions and Related Works}\label{s4c}

Comparing to existing state-of-the-art spatial graph convolution network models, the proposed BASGCN model has a number of advantages.

\textbf{First}, in order to transform the extracted multi-scale features from the graph convolution layers into fixed-sized representations, both the Neural Graph Fingerprint Network (NGFN) model~\cite{DBLP:conf/nips/DuvenaudMABHAA15} and the Diffusion Convolution Neural Network (DCNN) model~\cite{DBLP:conf/nips/AtwoodT16} sum up the extracted local-level vertex features as global-level graph features through a SumPooling layer. Although the fixed-sized features can be directly read by a classifier for classifications, it is difficult to capture local topological information residing on the local vertices through the global-level graph features. By contrast, the proposed BASGCN model focuses more on extracting local-level aligned grid vertex features through the proposed backtrackless spatial graph convolution operations (i.e., the in-spatial and the out-spatial graph convolution) on the aligned grid structures of graphs. Thus, the proposed BASGCN model can encapsulate richer local structural information than the NGFN and DCNN models associated with SumPooling.

\textbf{Second}, similar to the proposed BASGCN model, both the PATCHY-SAN based Graph Convolution Neural Network (PSGCNN) model~\cite{DBLP:conf/icml/NiepertAK16} and the Deep Graph Convolution Neural Network (DGCNN) model~\cite{DBLP:conf/aaai/ZhangCNC18} also need to form fixed-sized vertex grid structures for arbitrary-sized graphs. To achieve this, these models rearrange the vertex order of each graph structure, and preserve a specified number of vertices with higher ranks. Although, unify the number of vertices for different graphs, the discarded vertices may lead to significant information loss. By contrast, the associated aligned grid structures of the proposed BASGCN model can encapsulate all the original vertex features from the original graphs, thus the proposed BASGCN model constrains the shortcoming of information loss arising in the PSGCNN and DGCNN models. On the other hand, both the PSGCNN and DGCNN models tend to sort the vertices of each graph based on the local structural descriptor, ignoring consistent vertex correspondence information between different graphs. By contrast, the associated backtrackless aligned grid structure of the proposed BASGCN model is constructed through a transitive vertex alignment procedure. As a result, only the proposed BASGCN model can encapsulate the structural correspondence information between any pair of graph structures, i.e., the vertices on the same spatial position are also
transitively aligned to each other.

\textbf{Third}, as we have stated in Sec.\ref{SGCP}, both the backtrackless in-spatial and out-spatial graph convolution operations of the proposed BASGCN model are similar to performing standard fixed-sized convolution filters on standard grid structures. To further reveal this property, we utilize the in-spatial graph convolution operation as a typical instance and explain the convolution process one step further associated with Fig.\ref{f:alignment}. For the sample graph $G$ shown in Fig.\ref{f:alignment}, assume it has $5$ vertices following the fixed spatial vertex orders (positions) $v_1$, $v_2$, $v_3$, $v_4$ and $v_5$, $\bar{X}$ is the collection of its vertex feature vectors with $3$ feature channels, and $W^h$ is the $h$-th convolution filter with the filter size $5\times 1$ and the channel number $3$. Specifically, the procedure marked by the blue broken-line frame of Fig.\ref{f:alignment} indicates that performing the proposed in-spatial graph convolution operation associated with the in-adjacency matrix on the aligned vertex grid structure $\bar{X}$ can be seen as respectively performing the same $5\times 1$-sized convolution filter $W^h$ on five $5\times 1$-sized local-level in-neighborhood vertex grid structures included in the green broken-line frame. Here, each in-neighborhood vertex grid structure only encapsulates the original feature vectors of a root vertex as well as its in-adjacent vertices from $G$ (i.e., the vertices having directed edges to the root vertex), and all the vertices follow their original vertex spatial positions in $G$. For the non in-adjacent vertices, we assign dummy vertices (marked by the grey block) on the corresponding spatial positions of the in-neighborhood vertex grid structures, i.e., the elements of their feature vectors are all $0$. Since the five in-neighborhood vertex grid structures are arranged by the spatial orders of their root vertices from $G$, the vertically concatenation of these in-neighborhood vertex grid structures can be seen as a $25\times 1$-sized global-level grid structure $\bar{\mathbf{X}}_G$ of $G$. We observe that the process of the proposed in-spatial graph convolution operation on $\bar{X}$ is equivalent to sliding the $5\times 1$ fixed-sized convolution filter $W^h$ over $\bar{\mathbf{X}}_G$ with $5$-stride, i.e., this process is equivalent to sliding a standard classical convolution filter on standard grid structures. \textbf{As a result, the in-spatial graph convolution operation of the proposed BASGCN model is theoretically related to the classical convolution operation on standard grid structures, bridging the theoretical gap between traditional CNN models and the spatially-based GCN models}. Note that, we will obtain the same analysis result, if we utilize the out-spatial graph convolution operation as the typical instance.

\textbf{Fourth}, the above third observation indicates that both the in-spatial and out-spatial graph convolution operations are theoretically related to the classical convolution operation, and can assign each vertex a different weighted parameter. Thus, the proposed BASGCN model associated with the in-spatial and out-spatial graph convolution operations can adaptively discriminate the importance between specified in-neighboring or out-neighboring vertices during the convolution operation. By contrast, as we have stated in Sec.\ref{s2}, the existing spatial graph convolution operation of the DGCNN model only maps each vertex feature vector in $c$ dimensions to that in $c'$ dimensions, and all the vertices share the same trainable parameters. As a result, the DGCNN model has less ability to discriminate the importance of different vertices during the convolution operation.

\textbf{Fifth}, as we have stated in Sec.\ref{s2}, most existing spatially-based GCN models (e.g., the DGCNN, NGFN and DCNN, models) are theoretically related to the classical WL algorithm~\cite{shervashidze2010weisfeiler}. Similar to the WL algorithm, these GCN models suffer from tottering problem. This is because the associated graph convolution operations of these GCN models rely on the vertex feature information propagation through the undirected edges. As a result, they may propagate the feature information from the starting vertex to a vertex and then immediately propagate the information back to the starting vertex through the same undirected edge. By contrast, the proposed BASGCN model is defined based on the backtrackless aligned grid structure that corresponds to a directed line graph rather than an undirected graph. The associated in-spatial or out-spatial graph convolution operations cannot immediately propagate the vertex feature information against or along the directed edge. Thus, the proposed BASGCN model can significantly reduce the tottering problem arising in existing spatially-based GCN models.

\textbf{Finally}, similar to the proposed BASGCN model, the original ASGCN model~\cite{DBLP:conf/pkdd/Bai0BH19} cannot only reduce the information loss arising in most existing GCN models, but also bridge the theoretical gap between the traditional CNN models and spatially-based GCN models. This is because the ASGCN model is also based on the aligned grid structure computed based on the transitive vertex alignment method. However, similar to existing spatially-based GCN models, the original ASGCN model also suffers from the tottering problem. This is because, unlike the proposed BASGCN model, the associated spatial graph convolution operation of the ASGCN model is defined through the undirected grid vertex adjacency matrix (i.e., it is not defined on a backtrackless structure).
By contrast, the proposed BASGCN model is based on the backtrackless aligned grid structure, and can extract two kinds of multi-scale vertex features for each vertex though both the in-spatial and out-spatial graph convolution operations, thus reflecting richer graph characteristics than the original ASGCN model. Finally, note that, both the in-spatial and out-spatial graph convolution operations share the same trainable parameters. Thus, for the proposed BASGCN model, its associated in-spatial and out-spatial graph convolution operations are theoretically equivalent, if we replace the backtrackless grid structure as the backtracked grid structure used in the ASGCN model. Then, the BASGCN model will be as the same as the original ASGCN model, indicating that the proposed BASGCN model can generalize the original ASGCN model. As a result, \textbf{the proposed BASGCN model not only inherits all the advantages of the original ASGCN model, but also further generalizes the original model as a new backtrackless model to reduce the tottering problem and reflect richer graph characteristics}.

\section{Experiments}\label{s5}
In this section, we evaluate the performance of the proposed BASGCN model, and compare it to both state-of-the-art graph kernels and deep learning methods on graph classification problems. Specifically, the classification is evaluated with eight standard graph datasets that are abstracted from bioinformatics and social networks. Detailed statistics of these datasets are shown in Table.\ref{T:GraphInformation}.

\subsection{Comparisons on Graph Classification}
\begin{table*}
\vspace{-10pt}
\centering {
\scriptsize
\caption{Information of the Graph Datasets}\label{T:GraphInformation} \vspace{-10pt}
\begin{tabular}{|c||c||c||c||c||c||c||c||c|}
\hline
~Datasets ~        & ~MUTAG  ~ & ~PROTEINS~& ~D\&D~      & ~PTC~     & ~IMDB-B~   & ~IMDB-M~   & ~RED-B~    & ~COLLAB~  \\ \hline \hline
~Max \# vertices~  & ~$28$~    & ~$620$~   &  ~$5748$~   & ~$109$~   & ~$136$~    & ~$89$~     & ~$3783$~   & ~$492$~  \\ \hline
~Mean \# vertices~ & ~$17.93$~ & ~$39.06$~ &  ~$284.30$~ & ~$25.60$~ & ~$19.77$~  & ~$13.00$~  & ~$429.61$~ & ~$74.49$~  \\  \hline
~Mean \# edges~    & ~$19.79$~ & ~$72.82$~ &  ~$715.65$~ & ~$14.69$~ & ~$4914.99$~& ~$193.06$~ & ~$131.87$~ & ~$4914.99$~  \\  \hline
~\# graphs~        & ~$188$~   & ~$1113$~  &  ~$1178$~   & ~$344$~   & ~$1000$~   & ~$1500$~   & ~$2000$~   & ~$497.80$~  \\ \hline
~\# vertex labels~ & ~$7$~     & ~$61$~    &  ~$82$~     & ~$19$~    & ~$-$~      & ~$-$~      & ~$-$~      & ~$-$~  \\ \hline
~\# classes~       & ~$2$~     & ~$2$~     &  ~$2$~      & ~$2$~     & ~$2$~      & ~$3$~      & ~$2$~      & ~$3$~  \\ \hline
~Description~      & ~Bioinformatics~&~Bioinformatics~&~Bioinformatics~&~Bioinformatics~&~Social~&~Social~  & ~Social~  & ~Social~   \\ \hline

\end{tabular}
} \vspace{-10pt}
\end{table*}

\textbf{Experimental Setup:} We compare the performance of the proposed BASGCN model on graph classification applications with a) six alternative state-of-the-art graph kernels and b) twelve alternative state-of-the-art deep learning methods for graphs. Specifically, \textbf{the graph kernels include} 1) the Jensen-Tsallis q-difference kernel (JTQK) with $q=2$~\cite{DBLP:conf/pkdd/Bai0BH14}, 2) the Weisfeiler-Lehman subtree kernel (WLSK)~\cite{shervashidze2010weisfeiler}, 3) the shortest path graph kernel (SPGK) \cite{DBLP:conf/icdm/BorgwardtK05}, 4) the shortest path kernel based on core variants (CORE SP)~\cite{DBLP:conf/ijcai/NikolentzosMLV18}, 5) the random walk graph kernel (RWGK)~\cite{DBLP:conf/icml/KashimaTI03}, and 6) the graphlet count kernel (GK)~\cite{DBLP:journals/jmlr/ShervashidzeVPMB09}. On the other hand, \textbf{the deep learning methods include} 1) the deep graph convolutional neural network (DGCNN)~\cite{DBLP:conf/aaai/ZhangCNC18}, 2) the PATCHY-SAN based convolutional neural network for graphs (PSGCNN)~\cite{DBLP:conf/icml/NiepertAK16}, 3) the diffusion convolutional neural network (DCNN)~\cite{DBLP:conf/nips/AtwoodT16}, 4) the deep graphlet kernel (DGK)~\cite{DBLP:conf/kdd/YanardagV15}, 5) the graph capsule convolutional neural network (GCCNN)~\cite{DBLP:journals/corr/abs-1805-08090}, 6) the anonymous walk embeddings based on feature driven (AWE)~\cite{DBLP:conf/icml/IvanovB18}, 7) the edge-conditioned convolutional networks (ECC)~\cite{DBLP:conf/cvpr/SimonovskyK17}, 8) the high-order graph convolution network (HO-GCN)~\cite{DBLP:journals/corr/abs-1810-02244}, 9) the graph convolution network based on Differentiable Pooling (DiffPool)~\cite{DBLP:conf/nips/YingY0RHL18}, 10) the graph convolution network based on Self-Attention Pooling (SAGPool)~\cite{DBLP:conf/icml/LeeLK19}, 11) the graph convolutional network with EigenPooling (EigenPool)~\cite{DBLP:conf/icml/LeeLK19}, and 12) the degree-specific graph neural networks (DEMO-Net)~\cite{DBLP:journals/corr/abs-1906-02319}. Finally, to further demonstrate the advantages of the required backtrackless aligned grid structure, we also perform the proposed BASGCN model on the original un-backtrackless aligned grid structure (BASGCN(U)). As we have stated earlier, the proposed BASGCN(U) model corresponds to the original ASGCN model~\cite{DBLP:conf/pkdd/Bai0BH19}.

\begin{table*}
\vspace{-0pt}
\centering {
 \scriptsize
\caption{Classification Accuracy (In $\%$ $\pm$ Standard Error) for Comparisons with Graph Kernels.}\label{T:ClassificationGK}
\vspace{-10pt}
\begin{tabular}{|c|c|c|c|c|c|c|c|c|c|c|}

  \hline
 ~Datasets~& ~MUTAG  ~         & ~PROTEINS~       & ~D\&D~              & ~PTC~        & ~IMDB-B~       & ~IMDB-M~          & ~RED-B~  & ~COLLAB~\\ \hline \hline

 ~\textbf{BASGCN}~   & ~$\textbf{90.04}\pm0.82$~& ~$\textbf{76.05}\pm0.57$~& ~$\textbf{81.20}\pm0.99$~  & ~$\textbf{60.50}\pm0.77$  & ~$\textbf{74.00}\pm0.87$  & ~$50.43\pm.77$   & ~$\textbf{91.00}\pm0.25$ & ~$\textbf{79.60}\pm0.83$\\ 
 ~\textbf{BASGCN(U)}~   & ~$\textbf{89.70}\pm0.85$~& ~$\textbf{76.50}\pm0.59$~& ~$\textbf{80.40}\pm0.95$~  & ~$\textbf{61.42}\pm0.75$  & ~$\textbf{73.86}\pm0.92$  & ~$50.86\pm.85$   & ~$90.60\pm0.24$ & ~$\textbf{78.75}\pm0.79$\\ \hline

  ~JTQK~   & ~$85.50\pm0.55$~  & ~$72.86\pm0.41$~ &  ~$79.89\pm0.32$~   & ~$58.50\pm0.39$~ &~$72.45\pm0.81$~&  ~$50.33\pm0.49$~ & ~$77.60\pm0.35$  & ~$76.85\pm0.40$\\ \hline

  ~WLSK~   & ~$82.88\pm0.57$~  & ~$73.52\pm0.43$~ &  ~$79.78\pm0.36$~   & ~$58.26\pm0.47$~ &~$71.88\pm0.77$~&  ~$49.50\pm0.49$~ & ~$76.56\pm0.30$  & ~$77.39\pm0.35$\\   \hline

  ~SPGK~   & ~$83.38\pm0.81$~  & ~$75.10\pm0.50$~ &  ~$78.45\pm0.26$~   & ~$55.52\pm0.46$~ &~$71.26\pm1.04$~&  ~$\textbf{51.33}\pm0.57$~ & ~$84.20\pm0.70$ & ~$58.80\pm0.20$\\  \hline

  ~CORE SP~& ~$88.29\pm1.55$~  & ~$-$~            &  ~$77.30\pm0.80$~   & ~$59.06\pm0.93$~ &~$72.62\pm0.59$~&  ~$49.43\pm0.42$~ & ~$90.84\pm0.14$& ~$-$\\  \hline

  ~  GK~   & ~$81.66\pm2.11$~  & ~$71.67\pm0.55$~ &  ~$78.45\pm0.26$~   & ~$52.26\pm1.41$~ &~$65.87\pm0.98$~&  ~$45.42\pm0.87$~ & ~$77.34\pm0.18$& ~$72.83\pm0.28$\\ \hline

  ~RWGK~   & ~$80.77\pm0.72$~  &~$74.20\pm0.40$~  &  ~$71.70\pm0.47$~   & ~$55.91\pm0.37$~ &~$67.94\pm0.77$~&  ~$46.72\pm0.30$~ & ~$72.73\pm0.39$& ~$-$ \\ \hline

\end{tabular}
} \vspace{-10pt}
\end{table*}

\begin{table*}
\vspace{-0pt}
\centering {
 \scriptsize
\caption{Classification Accuracy (In $\%$ $\pm$ Standard Error) for Comparisons with Deep Learning Methods.}\label{T:ClassificationDL}
\vspace{-10pt}
\begin{tabular}{|c|c|c|c|c|c|c|c|c|c|}

  \hline
 ~Datasets~& ~MUTAG  ~       & ~PROTEINS~      & ~D\&D~            & ~PTC~        & ~IMDB-B~        & ~IMDB-M~         & ~RED-B~  & ~COLLAB~  \\ \hline \hline

 ~\textbf{BASGCN}~   & ~$\textbf{90.04}\pm0.82$~& ~$76.05\pm0.57$~& ~$\textbf{81.20}\pm0.99$~  & ~$60.50\pm0.77$  & ~$74.00\pm0.87$  & ~$50.43\pm.77$   & ~$\textbf{91.00}\pm0.25$ & ~$\textbf{79.60}\pm0.83$\\ 
 ~\textbf{BASGCN(U)}~   & ~$\textbf{89.70}\pm0.85$~& ~$\textbf{76.50}\pm0.59$~& ~$\textbf{80.40}\pm0.95$~  & ~$61.42\pm0.75$  & ~$73.86\pm0.92$  & ~$50.86\pm.85$   & ~$\textbf{90.60}\pm0.24$ & ~$\textbf{78.75}\pm0.79$\\ \hline

  ~DGCNN~  & ~$85.83\pm1.66$~& ~$75.54\pm0.94$~& ~$79.37\pm0.94$~  & ~$58.59\pm2.47$ & ~$70.03\pm0.86$ & ~$47.83\pm0.85$  & ~$76.02\pm1.73$& ~$73.76\pm0.49$\\ \hline

  ~PSGCNN~ & ~$88.95\pm4.37$~& ~$75.00\pm2.51$~& ~$76.27\pm2.64$~  & ~$62.29$        & ~$71.00\pm2.29$ & ~$45.23\pm2.84$  & ~$86.30\pm1.58$& ~$72.60\pm2.15$\\ \hline

  ~DCNN~   & ~$66.98$~       & ~$61.29\pm1.60$~& ~$58.09\pm0.53$~  & ~$58.09\pm0.53$ & ~$49.06\pm1.37$ & ~$33.49\pm1.42$  & ~$-$& ~$52.11\pm0.71$\\ \hline

  ~GCCNN~  & ~$-$~           & ~$76.40\pm4.71$~& ~$77.62\pm4.99$~  & ~$\textbf{66.01}\pm5.91$ & ~$71.69\pm3.40$ & ~$48.50\pm4.10$  & ~$87.61\pm2.51$& ~$77.71\pm2.51$\\ \hline

  ~DGK~    & ~$82.66\pm1.45$~& ~$71.68\pm0.50$~& ~$78.50\pm0.22$~  & ~$57.32\pm1.13$  & ~$66.96\pm0.56$ & ~$44.55\pm0.52$  & ~$78.30\pm0.30$& ~$73.09\pm0.25$\\ \hline

  ~AWE~    & ~$87.87\pm9.76$~& ~$-$~           & ~$71.51\pm4.02$~  & ~$-$             & ~$73.13\pm3.28$ & ~$\textbf{51.58}\pm4.66$  & ~$82.97\pm2.86$& ~$70.99\pm1.49$\\ \hline




  ~HO-GCN~    & ~$86.10$~& ~$-$~& ~$75.50$~  & ~$60.90$  & ~$\textbf{74.20}$ & ~$49.50$  & ~$-$& ~$-$\\ \hline

\end{tabular}
} \vspace{-10pt}
\end{table*}

%
%
%
%
%
%
%
%
%
%

For the evaluation, \textbf{ we employ the same network structure for the proposed BASGCN model on all graph datasets}. As we have stated earlier, the BASGCN model consists of two paralleling GCN models, i.e., the In-BASGCN network focusing on aggregating vertex features of itself and its in-neighboring vertices and the Out-BASGCN network focusing on aggregating vertex features of itself and its out-neighboring vertices, where both the networks share the same trainable parameters. Specifically, for either the IN-BASGCN or the Out-BASGCN network, we commence by setting the number of the prototype representations as $M=64$, because we observe that about $60\%$ to $70\%$ of the graphs have less than 64 vertices in our experiments. This can guarantee that the proposed model not only preserves all original vertices, but also retains the independent edge connections between vertices as much as possible. In other words, most edge connections between vertices will not be merged into one edge during the process of transforming each arbitrary sized graph into the fixed-sized grid structure. Moreover, for the IN-BASGCN or the Out-BASGCN network, we set the number of the proposed in-spatial or out-spatial graph convolution layers as $5$, and the number of the spatial graph convolutions in each layer as $32$. Based on Fig.\ref{f:QSGCNN} and Sec.\ref{s:arc}, we will get $6$ concatenated outputs after the In-BASGCN or the Out-BASGCN network, and we utilize a traditional one-dimensional CNN layer with the architecture as C$32$-P$2$-C$32$-P$2$-C$32$-F$128$ to further learn the extracted patterns, where C$k$ denotes a traditional convolutional layer with $k$ channels, P$k$ denotes a classical AvgPooling layer of size and stride $k$, and FC$k$ denotes a fully-connected layer consisting of $k$ hidden units. The filter size and stride of each C$k$ are all $5$ and $1$. With the six sets of extracted patterns after the CNN layers from the In-BASGCN or the Out-BASGCN network to hand, we concatenate and input them into a new fully-connected layer. For the In-BASGCN and the Out-BASGCN networks, we concatenate the features from their fully-connected layer and input the concatenated features to a Softmax layer with a dropout rate of $0.5$. We use the rectified linear units (ReLU) in both the graph convolution and the traditional convolution layer. The learning rate of the proposed model is $0.00005$ for all datasets. The only hyperparameters we optimized are the number of epochs and the batch size for the mini-batch gradient decent algorithm. To optimize the proposed BASGCN model, we use the Stochastic Gradient Descent with the Adam updating rules. Finally, note that, our model needs to construct the prototype representations to identify the transitive vertex alignment information over all graphs. In this evaluation we propose to compute the prototype representations from both the training and testing graphs. Thus, our model is an instance of transductive learning~\cite{DBLP:conf/uai/GammermanAV98}, where all graphs are used to compute the prototype representations but the class labels of the testing graphs are not used during the training process. For our model, we perform $10$-fold cross-validation to compute the classification accuracies, with nine folds for training and one fold for testing. For each dataset, we repeat the experiment 10 times and report the average classification accuracies and standard errors in Table.\ref{T:ClassificationGK}.

For the alternative graph kernels, we follow the parameter setting from their original papers. We perform $10$-fold cross-validation using the LIBSVM implementation of C-Support Vector Machines (C-SVM) and we compute the classification accuracies. We perform cross-validation on the training data to select the optimal parameters for each kernel and fold. We repeat the experiment 10 times for each kernel and dataset and we report the average classification accuracies and standard errors in Table.\ref{T:ClassificationGK}. Note that for some kernels we directly report the best results from the original corresponding papers, since the evaluation of these kernels followed the same setting of ours. For the alternative deep learning methods, we report the best results for the PSGCNN, DCNN, DGK models from their original papers, since these methods followed the same setting of the proposed model. For the AWE model, we report the classification accuracies of the feature-driven AWE, since the author have stated that this kind of AWE model can achieve competitive performance on label dataset. Moreover, note that the PSGCNN and ECC models can leverage additional edge features, most of the graph datasets and the alternative methods do not leverage edge features. Thus, we do not report the results associated with edge features in the evaluation. The classification accuracies and standard errors for each deep learning method are also shown in Table.\ref{T:ClassificationDL}. Finally, since the SAGPool, EigenPool, DEMO-Net models have not been evaluated on the social network datasets by the original authors, and ECC and the DiffPool models are only evaluated on one social network dataset (i.e., the COLLAB dataset) by the original author where the accuracies (67.79 and 75.48) are obviously lower than ours. For fair comparisons, we only report the accuracies of these models on the bioinformatics datasets in Table.\ref{T:ClassificationDLB}. 

\begin{table}
\vspace{-0pt}
\centering {
 \scriptsize
\caption{Classification Accuracy for Comparisons with Deep Learning Methods on Bioinformatics Datasets.}\label{T:ClassificationDLB}
\vspace{-10pt}
\begin{tabular}{|c|c|c|c|c|c|}

  \hline
 ~Datasets~    & ~MUTAG  ~       & ~PROTEINS~      & ~D\&D~            & ~PTC~     \\ \hline \hline

 ~\textbf{BASGCN}~      & ~$\textbf{90.04}$~& ~$76.05$~& ~$\textbf{81.20}$~  & ~$\textbf{60.50}$ \\ 
 ~\textbf{BASGCN(U)}~   & ~$\textbf{89.70}$~& ~$76.50$~& ~$80.40$~  & ~$\textbf{61.42}$ \\ \hline

  ~ECC~        & ~$76.11$~       &~$-$~            & ~$72.54$~         & ~$-$~                     \\ \hline

  ~DiffPool~   & ~$82.66$~ & ~$76.25$~       & ~$80.64$~         & ~$-$   \\ \hline

  ~SAGPool~    & ~$-$~& ~$71.86$~& ~$76.45$~  & ~$-$  \\ \hline

  ~EigenPool~  & ~$79.50$~& ~$\textbf{78.60}$~& ~$76.60$~  & ~$-$  \\ \hline

  ~DEMO-Net~    & ~$81.40$~& ~$-$~& ~$70.80$~  & ~$57.20$  \\ \hline

\end{tabular}
} \vspace{-10pt}
\end{table}
\textbf{Experimental Results and Discussions:} Table.\ref{T:ClassificationGK}, Table.\ref{T:ClassificationDL} and Table.\ref{T:ClassificationDLB} indicate that the proposed BASGCN model as well as its un-backtrackless version (i.e., the BASGCN(U) model) can significantly outperform both the remaining graph kernel methods and the remaining deep learning methods for graph classification. Specifically, for the alternative graph kernel methods, only the accuracy of the SPGN kernel on the IMDB-M dataset is a little higher than the proposed BASGCN and BASGCN(U) models. However, the proposed models are still competitive on the IMDB-M and RED-B datasets. On the other hand, for the alternative deep learning methods evaluated on both the bioinformatics and the social network datasets, only the accuracies of the GCCNN, HO-GCN and AWE models on the PTC, IMDB-M and IMDB-B datasets are higher than the proposed BASGCN and BASGCN(U) models. But the proposed models are still competitive on the IMDB-M and IMDB-B datasets. Moreover, for the alternative deep learning methods only evaluated on the social network datasets, only the accuracy of the EigenPool model on the PROTEINS dataset is higher than the proposed methods. Overall, the reasons for the effectiveness are fourfold. 

First, all the graph kernels for comparisons fall into the instances of R-convolution kernels. Since these kernels only focus on the isomorphism measure between any pair of substructures without considering the structural location within the global graph structure. These kernel methods tend to ignore the structure correspondence information between the substructures. By contrast, the proposed BASGCN and the BASGCN(U) models can incorporate the transitive alignment information between the vertices over all graphs, through the associated aligned grid structure. As a result, the proposed model can better reflect the precise characteristics of graphs. On the other hand, it is shown that the C-SVM classifier associated with graph kernels are instances of shallow learning methods~\cite{DBLP:conf/icassp/ZhangLYG15}. By contrast, the proposed model can provide an end-to-end deep learning architecture, and thus better learn graph characteristics.

Second, as instances of spatially-based GCN models, the trainable parameters of the DGCNN, ECC, DCNN, HO-GCN, DEMO-Net, SAGPool and EigenPool models are shared for each vertex. Thus, these models cannot directly influence the aggregation process of the vertex features. By contrast, the required backtrackless graph convolution operation of the proposed BASGCN model is theoretically related to the classical convolution operation on standard grid structures and can adaptively discriminate the importance between specified vertices. Moreover, since these spatially-based GCN model are also theoretically related to the classical WL algorithm that suffers from the tottering problem, they may generate redundant information in the process of graph convolution. By contrast, the proposed BASGCN model can significantly reduce the tottering problem through the associated backtrackless graph convolution operation.

Third, similar to the R-convolution graph kernels, the DGCNN, PSGCNN, DCNN, GCCNN, DGK, AWE, HO-GCN, ECC, SAGPool, EigenPool and DEMO-Net models cannot integrate the correspondence information between graphs into the learning architecture. Especially, the PSGCNN, ECC and DGCNN models need to reorder the vertices to construct fix-sized grid structures and some vertices may be discarded in the process, resulting in significant information loss. By contrast, the associated (un)backtrackless aligned vertex grid structures of the proposed BASGCN and BASGCN(U) models can preserve more information from original graphs, reducing the problem of information loss. Moreover, although the DiffPool model can also integrate the vertex alignment information in the Differentiable Pooling operation to reflect correspondence information between graphs. Unfortunately, its associated graph convolution operation follows the form of spatially-based GCN models. As a result, the DiffPool model is also an instance of the WL analogous GCN models and suffers from the tottering problem.

Fourth, unlike the proposed model, the DCNN, ECC, and DEMO-Net model need to sum up the extracted local-level vertex features as global-level graph features. By contrast, the proposed BASGCN and BASGCN(U) models focus more on local structures and can learn richer multi-scale local-level vertex features.

Finally, note that, the proposed BASGCN model associated with the backtrackless aligned grid structure can outperform that associated with the un-backtrackless aligned grid structure (i.e., the BASGCN(U) model) on most datasets. Although the accuracies of the BASGCN model are lower than that of the BASGCN(U) model on the PROTEINS, PTC and IMDB-M datasets, the BASGCN model is still competitive. As we have stated earlier, the BASGCN(U) corresponds to the original ASGCN model~\cite{DBLP:conf/pkdd/Bai0BH19} that also suffers the tottering problem. This indicates that the proposed model not only inherit the advantages of the original ASGCN model, but also generalizes the original model as a new backtrackless model to reduce the tottering problem and reflect richer graph characteristics.

\subsection{The Efficiency of the Proposed Model}

In this subsection, we empirically evaluate the computational efficiency of the proposed BASGCN model, and compare it to the popular WLSK kernel~\cite{shervashidze2010weisfeiler} that is one of the most efficient graph kernels. Moreover, we compare the runtime of the proposed BASGCN model and the WLSK kernel on the RED-B benchmark dataset. The reason of choosing this dataset is that the average size of its graphs is the largest  in our experimental evaluation. Specifically, for the RED-B dataset, the WLSK kernel takes $2,170$ seconds to compute the kernel matrix, and another $837$ seconds to train the C-SVM associated with the kernel matrix for one round of 10-fold cross validation. For the proposed BASGCN model, computing the fixed-sized grid structures takes $3,627$ seconds, and another $167$ seconds to train the BASGCN model. Note that the training time of the proposed BASGCN model relates to the choice of the epoch number, and we set the epoch number as $10$. This is because the proposed BASGCN model can already obtain better classification accuracy than the WLSK kernel under this setup. Thus, the overall runtimes for the proposed BASGCN model and the WLSK kernel are $3,794$ seconds versus $3,007$ seconds. Although the runtime of the proposed BASGCN model is slightly higher than that of the WLSK kernel, the computational efficiency of the proposed model is still competitive to the WLSK kernel. Moreover, the proposed model can significantly outperform the WLSK kernel in terms of classification accuracy. In summary, compared to the efficient WLSK kernel, the proposed ASGCN model has a better tradeoff between classification accuracy and the computational efficiency.

\subsection{Other Performance Evaluation of the Proposed Model}
\begin{figure}
\centering
\subfigure{\includegraphics[width=1\linewidth]{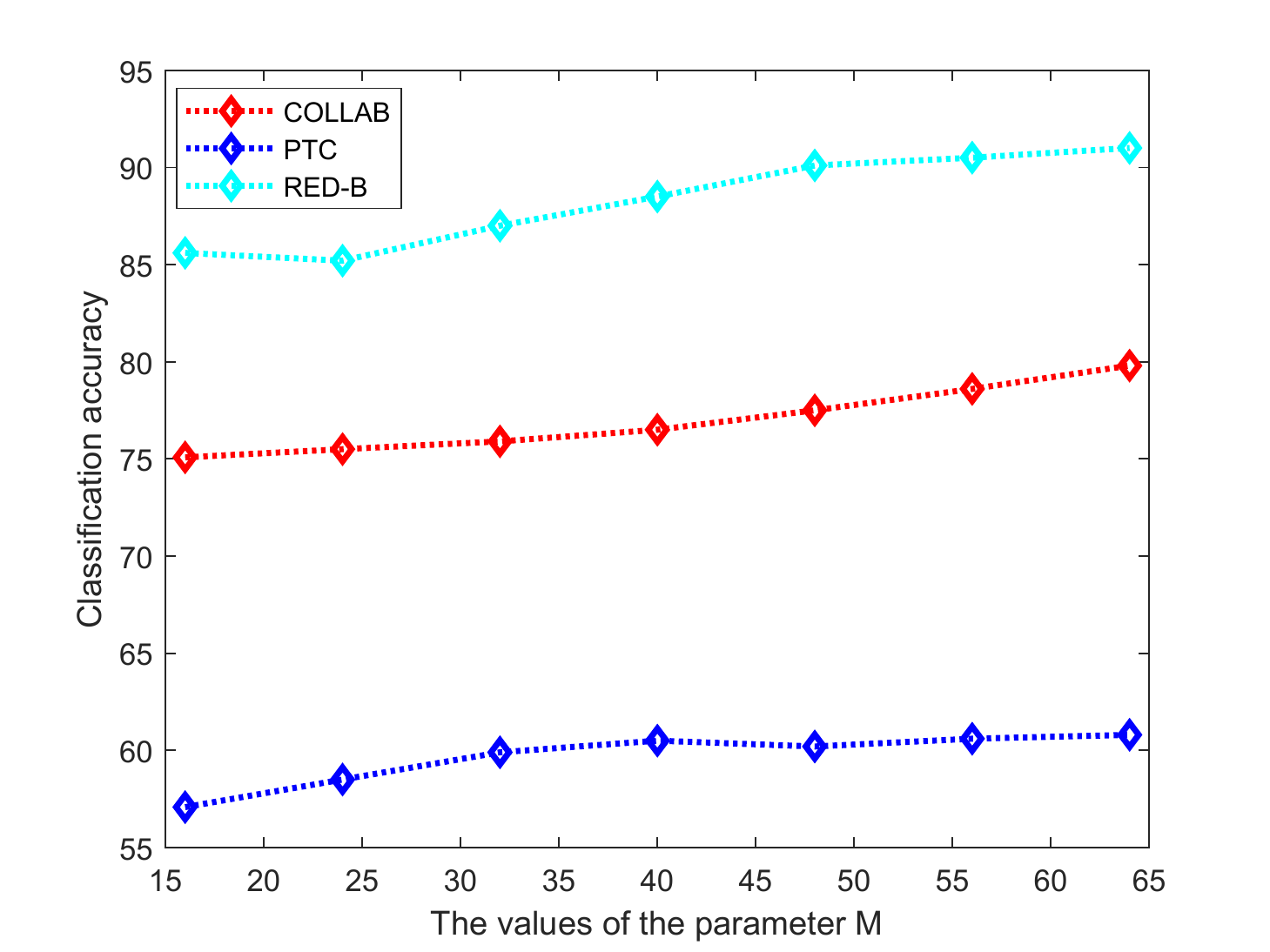}}
\vspace{-10pt}
\caption{Accuracy vs the parameter M.}\label{aparametereva}
\vspace{-15pt}
\end{figure}

In this subsection, to indicate the performance of the proposed model one step further, we evaluate how the selection of the parameter $M$ influences the classification performance of the proposed BASGCN model on the COLLAB, PTC and RED-B datasets. The reason of choosing these three datasets for this evaluation is due to their representativeness in terms of different levels of graphs size and number. In fact, we will observe similar phenomenon on the remaining datasets. Specifically, we vary the parameter $M$ from 16 to 64 (with steps of size 8), and Figure \ref{aparametereva} exhibits how the classification accuracy of the proposed BASGCN model varies with increasing values of $M$. Through Fig.\ref{aparametereva}, we observe that the classification accuracies of the proposed model gradually improve and tend to be stable with increasing $M$.

\section{Conclusions}\label{s6}

In this paper, we have developed a new spatially-based GCN model, namely the Backtrackless Aligned-Spatial Graph Convolutional Network (BASGCN) model, to learn effective features for graph classification. This model is based on transforming the arbitrary-sized graphs into fixed-sized backtrackless aligned grid structures, and performing a new backtrackless spatial graph convolution operation on the grid structures. Unlike most existing spatially-based GCN models, the proposed BASGCN model cannot only adaptively discriminate the importance between specified vertices during the process of the spatial graph convolution operation, but also significantly reduce the notorious tottering problem of existing spatially-based GCNs related to the Weisfeiler-Lehman algorithm. Experiments on standard graph datasets demonstrate the effectiveness of the proposed model.

In this work, we adopted the consistent network architecture as well as the same parameter setting (excluding the numbers of the epoch and mini-batch) for all datasets. Our future works will aim to learn the optimal structure and parameter setting for each individual dataset, which should in turn improve the classification performance.


\balance


\bibliographystyle{IEEEtran}
\bibliography{example_paper}

\end{document}